%% file: main.tex
\input{config/paperMetaData.tex}

\input{config/journalConfigs/laPreprint.tex}
\input{config/journalConfigs/laPreprint_customization.tex}
\input{config/commonConfigStuff.tex}
\input{config/bibConfigs.tex}

\newcommand{\fastDsaTermFull}{fast dynamical similarity analysis}
\newcommand{\fastDsaTermAbbriv}{fastDSA}
\newcommand{\fastDsaTermAbbrivPlu}{fastDSA's}
\newcommand{\fastDsaTermAbbrivCap}{FastDSA}

\begin{document}
\input{config/journalConfigs/laPreprint_opening.tex}

\begin{abstract}
To understand how nonlinear dynamical systems process information (e.g., artificial neural networks and neural circuits), it is essential to compare the underlying dynamics of diverse systems at scales (e.g., a large pool of neural networks with diverse architectures and large-scale recordings from neural circuits).
Although a substantial body of similarity metrics has been developed, existing approaches remain inadequate for large-scale comparisons of systems.
On one hand, geometric similarity methods are computationally efficient and allow large-scale comparisons; however, they are inaccurate as they do not capture the \emph{governing dynamics}, which is essential to understand these nonlinear dynamical systems and the computations they implement. 
On the other hand, conventional methods to compare the \emph{similarity of dynamics} are faithful to governing dynamics but often impractical due to their computational cost.    
%
Here, we bring the best of both worlds, the computational efficiency of geometric tools and the reliability of dynamic similarity methods, enabling us to compare diverse nonlinear dynamical systems at large scales. 
We introduce  \emph{\fastDsaTermFull} (\fastDsaTermAbbriv) as a computationally efficient and accurate (dis)similarity metric between two nonlinear dynamical systems. 
\fastDsaTermAbbrivCap\ is built based on modern computational methods: methods from random matrix theory for the detection of the optimal rank of a dynamical system, 
developing multiple novel optimization pipelines to align the flow field of dynamical systems, and powerful Koopman embeddings.
We demonstrate that across multiple benchmark nonlinear systems and recurrent network models, \fastDsaTermAbbriv\ is insensitive to arbitrary coordinate choices, but sensitive to real differences in the systems' dynamics (how the systems evolve over time), which geometric methods may miss and are computationally expensive with traditional dynamical methods. 
To the best of our knowledge, \fastDsaTermAbbriv\ is the fastest yet accurate method to compare the (dis)similarity of nonlinear dynamical systems.
\fastDsaTermAbbrivCap\ allows large-scale and statistical analyses of a wide range of nonlinear dynamical systems, consequently, extending the practical scope of dynamical similarity analysis.

\vspace{6pt}
\textcolor{gray}{\textbf{Keywords:}}\\
Nonlinear dynamics, 
Dynamical systems,
Similarity analysis, 
Dynamical systems analysis at scale,
Neural networks, 
Neural computations,
Computations through dynamics

\end{abstract}

\section{Main}
Nonlinear dynamical systems serve as a robust foundation for diverse information processing architectures \citep{sussillo2014_neural, kim2023neural, driscoll2024flexible, Oby2025DynamicalConstraints, vyas2020computation}. 
In particular, neural networks, whether biological or artificial, encode, store, and transform information through their complex dynamics \citep{mante2013context, Oby2025DynamicalConstraints, vyas2020computation, luo2025transitions, pagan2025individual, perich2025neural, chaudhuri2019intrinsic, driscoll2024flexible, jensen2024recurrent, ji2025discovering, genkin2025dynamics, tolmachev2025single, sohn2019bayesian, safavi2024signatures, stringer2024analysis}. Comparing the dynamics of diverse systems is fundamental for uncovering their information processing mechanisms \citep{kriegeskorte2008representational, mante2013context, chaudhuri2016computational, raghu2017svcca, schrimpf2018brain, chaudhuri2019intrinsic, schaeffer2020reverse, vyas2020computation, williams2021generalized, duong2022representational, zeraati2024neural, voges2024decomposing, meilua2024manifold}. Methods that compare dynamics between systems, such as real-world data and models, allow us to gain an algorithmic insight into neural computations \citep{voges2024decomposing} and ultimately a better understanding of the machinery supporting cognitive computations \citep{wang2012neural, zoltowski2020general, cueva2020low, shine2021author, safavi2022multistability, wang2022theory, beiran2023parametric, whyte2024thalamic, munn2024multiscale, wang2024bifurcation, voigts2025spatial, chandra2025episodic}. 
Thus, tools to compare the dynamics of diverse neural networks at scales are essential to understand how they process information.  

Although numerous similarity metrics have been proposed to compare neural networks, they are not practical for studying \emph{the dynamics of neural networks with diverse characteristics on large scales} --- which is crucial to understand their dynamical principles serving information processing \citep{driscoll2024flexible, Oby2025DynamicalConstraints, vyas2020computation}. 
A large subclass of similarity measures primarily focuses on the geometry of the data, or on multivariate statistics \citep{schrimpf2018brain, kriegeskorte2008representational, raghu2017svcca, williams2021generalized, duong2022representational, barbosa2025quantifying}. 
While these methods are computationally efficient, they often struggle to capture the dynamics governing the neural system, which is essential to understand how they realize computations \citep{ostrow2023beyond, huang2025inputdsa}. 
Thus, comparing the underlying neural dynamics in these systems was suggested as a more faithful approach to compare neural information processing systems, an approach that has recently been gaining currency \citep{ostrow2023beyond, eisen2024propofol, ostrow2024delay, lipshutz2024disentangling, chen2024dform, guilhot2024dynamical, codol2024brain, bolager2024gradient, huang2024measuring, kujur2025transient, xu2025nn, huang2025inputdsa}. 
Unlike geometrical methods that focus on geometrical and/or statistical similarities, methods for dynamic similarity analysis \citep{eisen2024propofol, ostrow2024delay, guilhot2024dynamical, codol2024brain, bolager2024gradient, huang2024measuring, kujur2025transient, xu2025nn, huang2025inputdsa} emphasize the fundamental properties of systems' dynamics by leveraging insights from the theory of dynamical systems \citep{thompson2002nonlinear, strogatz2024}. 
This new class of (dynamic) similarity metrics is conceptually appealing; however, their high computational cost makes them impractical for large-scale analysis. 
Specifically, it is too expensive to deploy across extensive pools of networks featuring diverse architectures, input regimes, and learning curricula.
The ideal solution combines the best of both approaches: the computational efficiency of geometrical methods and the accuracy of dynamic similarity metrics.

In this study, we introduce \emph{\fastDsaTermFull} (\fastDsaTermAbbriv) that has the advantages of both worlds: 
accuracy and mathematical properties of methods based on dynamic similarity analysis
and the computational efficiency of geometrical methods. 
\fastDsaTermAbbrivCap\ uses the principles of Koopman theory to capture the most essential aspects of nonlinear dynamical systems \citep[similar to, ][]{eisen2024propofol, ostrow2024delay, guilhot2024dynamical, codol2024brain, huang2024measuring, kujur2025transient, xu2025nn, huang2025inputdsa} and targets the similarity of dynamical systems by estimating their topological conjugacy. 
It does so by embedding nonlinear dynamics into a globally linear space using Koopman theory \cite{snyder2021koopman}. 
To make this comparison practical at scale, \fastDsaTermAbbriv\ introduces automated model-order (rank) selection for Hankel--Koopman embeddings and efficient optimization schemes for near-orthogonal alignment, substantially reducing the computational bottlenecks of operator-alignment dynamical similarity.
As a result, it can distinguish systems with similar state-space geometry but different dynamics, and it can identify dynamical similarity across systems with distinct spatial configurations and/or architectures.

Through a set of diverse analyses, on simulated nonlinear dynamical systems and recurrent neural networks, we demonstrate that \fastDsaTermAbbriv\ successfully combines the strength of both worlds (computational efficacy and reliable dynamic similarity assessment) for comparing the (dis)similarity of nonlinear dynamical systems. 
First, using a ring attractor network, we show that \fastDsaTermAbbriv\ maintains strict invariance to topology-preserving geometric deformations. 
Second, by parametrically transforming the ring into a line attractor, we confirm that \fastDsaTermAbbriv\ accurately detects genuine topological shifts in the governing dynamics. Finally, we compare \fastDsaTermAbbriv\ to fast baseline similarity measures. This analysis tests two practical desiderata: robustness to noise and sensitivity to fine-grained differences in dynamical mechanisms.
Notably, we demonstrate in all scenarios we evaluate \fastDsaTermAbbriv\ was computationally more efficient than conventional dynamical similarity metrics and more accurate than geometrical methods.
Altogether, we demonstrate \fastDsaTermAbbriv\ is a computationally efficient, mathematically solid, and accurate framework to compare dynamics across nonlinear dynamical systems (in particular neural networks).  
Most importantly, \fastDsaTermAbbriv\ enables large-scale analysis of nonlinear dynamical systems across models, training conditions, and experimental datasets.

\section{Results}

\emph{\fastDsaTermAbbrivCap}\ is an efficient and accurate method for estimating dynamic (dis)similarity between neural networks, neuronal circuits, or data with a model. 
In \fastDsaTermAbbriv, we integrate concepts from random matrix theory to identify a low-rank approximation of the system's dynamics, and develop new optimization procedures for aligning vector fields.
We introduce and compare three variants of our \fastDsaTermAbbriv\ (each based on a different optimization procedure), and show that all variants preserve the accuracy and mathematical properties of conventional dynamical similarity methods \citep{ostrow2023beyond} while substantially enhancing computational efficiency.
We demonstrate the capacity of the family of \fastDsaTermAbbriv\ methods with several numerical experiments with systematic manipulation of the governing dynamics in neural networks. 

\subsection{A scalable Koopman operator alignment for dynamical similarity} 
\label{sec:accel-dsa}

Our goal is to quantify dynamical similarity by comparing \emph{governing dynamics} rather than the geometry of data (e.g., statistical similarities).
We do so by constructing Koopman-based linear representations of each system and then measuring similarity through an \emph{operator-alignment}:
if two systems implement equivalent dynamics up to a change of coordinates, their induced operators can be aligned by an (approximately) orthogonal similarity transform, and if different, they will be proportionally further away from an orthogonal similarity transform. 
We formalize as follows.
Let \(x\) and \(y\) be two dynamical systems:
\begin{equation}
\dot{x} = f(x, t), \quad x \in \mathbb{R}^n, \quad \dot{y} = g(y, t), \quad y \in \mathbb{R}^m\,.
\end{equation}
Then, data from each system can be structured into Hankel matrices ($H_{x}$, $H_{y}$). The delay embedding effectively lifts the nonlinear dynamics into a higher-dimensional space where, according to Koopman theory (in the limit of infinite-dimensional embeddings), they can be approximated as linear \citep{arbabi2017ergodic,brunton2021modernkoopmantheorydynamical}:
\begin{equation}
V'_{x,r} = A_x V_{x,r}, \quad \text{where } H_x^T = U_x \Sigma_x V_x^T, \quad V_{x,r} = V_x[:, 1:r]\,,
\label{eq:dmdRank}
\end{equation} 
where $H_x^T = U_x \Sigma_x V_x^T$ is the singular value decomposition (SVD) of the transposed Hankel matrix,  
$H$; $\Sigma$, and $V$ are the standard SVD components; 
$r$ denotes the chosen truncation rank; 
$V_x[:,1:r]$ consists of the first \(r\) columns of \(V_x\); 
$A_x$ is the matrix that provides the approximate linear dynamics in the reduced subspace, 
and $V'_{x,r}$ represents the next predicted step in this coordinate space.
The resulting data matrices represent the approximated linear dynamics. This numerical methodology, also known as the Dynamic Mode Decomposition (DMD), serves as a practical method for analyzing complex nonlinear dynamics through a linear approximation \citep{brunton2017chaos}. 
The DMD matrix $A$ is typically decomposed into eigenvectors and eigenvalues, which capture core dynamic properties of the data. Theoretically, the DMD seeks a finite approximate of the Koopman operator \citep{arbabi2017ergodic,brunton2021modernkoopmantheorydynamical} that linearize the evolution of \textit{functions} on the state space. 
The DMD identifies a linear operator that can estimate the system's dynamics in the next time step \( t+\Delta t \),
\(
X_{(t + \Delta t)} = A X_{(t)}
\).
In previous studies \citep{arbabi2017ergodic, brunton2017chaos, ostrow2023beyond, kamiya2024koopman} the lifting step has often been instantiated using delay-coordinate (Hankel) embeddings followed by DMD/Hankel-DMD (HAVOK) to obtain a finite-dimensional Koopman approximation. Here, we adopt the same practical Koopman-embedding strategy; however, we emphasize that the broader operator-alignment view \citep[that is not limited to this choice, kernel methods, neural encoders, or alternative Koopman approximations can likewise map nonlinear dynamics into a representation where linear operator comparisons are meaningful; also see, ][]{ostrow2026gdsacosyne}.
On the next step, the linear dynamics approximations $A_{x}, A_{y}$, are compared using a modified Procrustes analysis tailored specifically for vector fields (Procrustes Analysis over Vector Fields, PAVF):
\begin{equation}
d(A_x, A_y) = \min_{C \in O(n)} ||A_x - C A_y C^{-1}||_{F}\,,
\label{eq:dsaEucDistance}
\end{equation}
where \(C\) is the transformation that best aligns \(A_x\) and \(A_y\),  $O(n)$ is the group of n-dimensional orthonormal matrices $(C^T C = I)$, $F$ denotes the Frobenius norm, and $X$, $Y$ are two given
data matrices. This metric measures the similarity between each system's dynamics, invariant to orthogonal transformations. 
An angular form of this metric can also be utilized, which can be computed as:
\begin{equation}
d(A_x, A_y) = \min_{C \in O(n)} \arccos\left(\frac{A_x \cdot (C A_y C^{-1})}{|A_x|_F |A_y|_F}\right)\,.
\end{equation}
This provides a principled metric to assess dynamical equivalence beyond geometric (dis)similarities.

\fastDsaTermAbbrivCap\ builds on well-established operator-alignment approaches \citep[e.g., ][]{ostrow2023beyond, chen2024dform, redman2024identifying, zhang2025koopstd, kamiya2024koopman}, but removes two bottlenecks that have limited existing methods to modest problem sizes:
(1) Calculating the approximate optimal rank automatically (using methods based on random matrix theory). 
(2) Incorporating novel optimization constraints for PAVF to reduce computational overhead  
(see \autoref{fig:fDSAshematic} for a schematic illustration of the family of \fastDsaTermAbbriv methods and their comparison; and section \nameref{sec:fast-optz-dsa} and \autoref{fig:AlgoritmStructure} for further elaboration).

\input{fig0.tex}

Building on prior operator-alignment formulations, we propose three alignment optimizers that retain the same objective and invariance guarantees while making the constrained alignment problem computationally tractable in large-sample settings.
These provide alternative ways to find the similarity transform for the orthogonal group \(O(n)\) (or \(SO(n)\)).
Given two matrices \(A_x\) and \(A_y\), 
we seek a matrix \(C\) such that
\begin{equation}
\label{eq:lossEuc}
    \min_{C\in O(n)} \; f(C),
    \qquad
    f(C) \;=\; \bigl\|A_x - C A_y C^\top\bigr\|_F^2\,.
\end{equation}
Here, similar to \autoref{eq:dsaEucDistance}, \(C\) is the transformation that best aligns $A_y$ with $A_x$ subject to the orthogonality constraint
\begin{equation}
    C^\top C = I, \qquad \text{(orthogonality condition for } O(n))\,.
    \label{eq:orthogonality}
\end{equation}
In prior operator-alignment formulations \citep[e.g., in][]{ostrow2023beyond}, the orthogonality constraint is typically reimposed after each update step, which can dominate runtime and hinder scalability.
\fastDsaTermAbbrivCap\ addresses this bottleneck by supporting three alignment solvers that optimize the same operator-alignment objective while handling near-orthogonality more efficiently:
(1) \fastDsaTermAbbriv\ with a regularized optimization (including a penalty term in the optimization procedure (we will refer to as \emph{RO-\fastDsaTermAbbriv}). 
(2) \fastDsaTermAbbriv\ that is optimized based on-manifold Riemannian method \citep{ablin2022fast} that we will refer to as \emph{Rim-\fastDsaTermAbbriv}).
(3) We use the Landing algorithm \citep{ablin2022fast} to optimize for the alignment problem of DSA (we will refer to it as \emph{Land-\fastDsaTermAbbriv}).
See section \nameref{sec:fast-optz-dsa} for the details of each optimization variant.
In the next sections, we demonstrate the practical implications of these novel optimizations along with other key components of \fastDsaTermAbbriv.

\subsection{Automatic estimation of optimal rank of dynamical systems}
\label{sec:supp-SVHT}

One of the main challenges of the DMD is to determine the appropriate rank for the SVD that is used in the linear representation of the system (also, see, \autoref{eq:dmdRank}). Selecting a rank that is too high can introduce spurious modes into the comparison, while a rank that is too low may cause the loss of essential information.
Furthermore, using an optimal rank can also potentially act as denoising, which facilitates estimation, given that noise components typically correspond to smaller singular values. 
Several methods have been proposed for automated and statistically-grounded rank selection \citep{nadlerDistributionRatioLargest2011,kritchmanNonParametricDetectionNumber2009,kritchmanDeterminingNumberComponents2008,veraartDenoisingDiffusionMRI2016,safaviUnivariateMultivariateCoupling2021,safaviUncoveringOrganizationNeural2023}. One notable approach is the Optimal Singular Value Hard Thresholding \citep[SVHT, ][]{gavish2014optimal}, 
which forms the basis of our method for estimating the optimal rank of dynamical systems and has a solid statistical foundation based on random matrix theory \citep{plerouRandomMatrixApproach2002,paulRandomMatrixTheory2014,leeRandomMatrixTheory2014,izenmanRandomMatrixTheory,karouiSpectrumEstimationLarge2008}.

The optimal hard threshold for an $m \times n$ data matrix $X$ observed in white noise of known variance $\sigma$ is given by:
\begin{equation}
    \tau^* = \lambda^* (\beta) \sqrt{n} \sigma\,,
\end{equation}
where $\beta = m/n$ and $\lambda^*(\beta)$ is a coefficient that depends on the aspect ratio of the data matrix. When the noise level is unknown, an adaptive threshold can be used based on the median empirical singular value:
\begin{equation}
    \tau^* = \omega(\beta) \cdot x_{\text{med}}\,,
\end{equation}
where $x_{\text{med}}$ is the median singular value of $X$, and $\omega(\beta)$ is a precomputed constant \cite[]{gavish2014optimal}.

First, to demonstrate the capability of SVHT for accurate rank detection in basic settings (e.g., determining the rank based on the latent dynamics), we employed synthetic data generated from the Lorenz attractor, a canonical example of a low-dimensional chaotic system with a priori known latent dimensionality (rank = 3).
The Lorenz attractor dynamics is described by the following system of ordinary differential equations:
\begin{align}
\dot{x} &= \sigma(y - x), \\
\dot{y} &= x(\rho - z) - y, \\
\dot{z} &= xy - \beta z\,,
\end{align}
with standard chaotic parameters set as $\sigma = 10$, $\rho = 28$, and $\beta = \frac{8}{3}$. The true latent state at any time is represented by the three-dimensional vector $(x, y, z)$, confirming an intrinsic rank of 3 (\autoref{fig:Figure2_new}a, top-left).
%
Thus, we simulated Lorenz trajectories by initializing the system at $[x_0, y_0, z_0] = [0, 1, 1.05]$ and numerically integrating over time. 
This process generates a multivariate timeseries with a known low-dimensional latent dynamics (with three dimensions and length $T$, representing the number of time steps).
Next, to simulate observational data with higher dimensionality, a random projection matrix $W \in \mathbb{R}^{3 \times D}$ was constructed, where $D$ is the number of observed dimensions (in this case 128). 
Application of the projector operator $W$ leads to a new higher-dimensional (\(D\)-dimensional) data matrix that retained an approximate intrinsic rank of 3, reflecting its low-dimensional origin \citep[similar to,][; see \autoref{fig:Figure2_new}a, top-right]{kapoor2024latent}.
We then demonstrate the effectiveness of our method for rank detection by testing it on dynamics with known rank. 

To further assess SVHT as a method for identifying optimal rank, we compared its rank selection accuracy against traditional information criteria such as the Akaike Information Criterion (AIC), Bayesian Information Criterion (BIC), and the corrected AIC \citep[AICc, ][see \autoref{fig:Figure2_new}a, bottom-left]{hurvich1993corrected}.
The conventional formulations of AIC and BIC were used for estimation here:
\begin{align}
\text{AIC}_r &= 2k_r + N_{\text{obs}} \ln\left(\frac{RSS_r}{N_{\text{obs}}}\right), \\
\text{BIC}_r &= \ln(N_{\text{obs}}) k_r + N_{\text{obs}} \ln\left(\frac{RSS_r}{N_{\text{obs}}}\right)\,,
\end{align}
where $RSS_r$ is the residual sum of squares computed from the top $r$ singular modes, $N_{\text{obs}} = n_{\text{time}} \times n_{\text{neurons}}$ represents the total number of observations, and $k_r = r(n_{\text{time}} + n_{\text{neurons}} - r)$ denotes the number of parameters for the rank-$r$ model.
Assuming i.i.d. Gaussian residuals with unknown variance $\sigma^2$, the log-likelihood under a rank-$r$ model is:
\begin{equation}
\ln L_r = -\frac{n}{2} \ln\left(2\pi\sigma^2\right) - \frac{RSS_r}{2\sigma^2}\,.
\end{equation}
Using the maximum likelihood estimate $\hat{\sigma}^2 = \frac{RSS_r}{n}$, this simplifies to:
\begin{equation}
-2\ln L_r = n \ln\left(\frac{RSS_r}{n}\right) + \text{constant in } r\,,
\end{equation}
For small sample sizes, we also computed the corrected AIC (AICc) given by:
\begin{equation}
\text{AICc} = \text{AIC}_r + \frac{2k_r(k_r + 1)}{N_{\text{obs}} - k_r - 1}\,.
\end{equation}

We also demonstrate the  SVHT robustness for rank detection in the presence of noise.
This capacity stems from its foundation in random matrix theory that exploit the known reflection of noise statistics in the spectrum of singular values \citep{nadlerDistributionRatioLargest2011,kritchmanNonParametricDetectionNumber2009,kritchmanDeterminingNumberComponents2008,veraartDenoisingDiffusionMRI2016,safaviUnivariateMultivariateCoupling2021, safaviUncoveringOrganizationNeural2023}. 
Thus, we used varying levels of noise to the projected data to assess the robustness of the SVHT-based method for determining the optimal rank. 
SVHT consistently identified the correct rank of the latent dynamics (rank 3) in moderately noisy conditions (\autoref{fig:Figure2_new}a, bottom-right).

To further assess SVHT in a practical forecasting context (which is central for our purpose, dynamics similarity analysis), we examined its utility in DMD-based reconstruction of time series data.
Since the SVHT was originally developed for denoising low-rank matrices, it is unclear whether this approach would also be applicable for selecting the optimally predicting rank in a Hankel matrix, as this is a different objective. 
We evaluate this and demonstrate that the SVHT is able to approximate the elbow of two performance measures of the DMD (\autoref{fig:Figure2_new}c-d). 
Given the observed data matrix $X$ consisting of a multivariate time-series, we construct its Hankel embedding (noted by $H$, \autoref{eq:dmdRank}). 
The rank of $H$ is then estimated using optimal SVHT, to focus primarily on the most informative singular values:
\begin{equation}
    r^* = \max \left( \hat{r}_1, \hat{r}_2 \right)\,,
\end{equation}
where $\hat{r}_1$ and $\hat{r}_2$ are the optimal ranks esimated for two different embeddings. The automatic choice of the best low-rank estimate facilitates the capture of the most significant dynamical structures while maintaining computational efficiency. 
%
Thus, this procedure refines rank determination by computing the optimal rank independently for both Hankel-embedded matrices (\autoref{eq:dmdRank}). To identify the optimal rank for both matrices, we compare the estimated ranks and ultimately select the larger one (see section \nameref{sec:supp-minmax} and \autoref{fig:minOrmax} for more details on the rationale behind this choice). 

We began by identifying optimal embedding parameters for the Lorenz attractor data and applied DMD after embedding. We then introduced two types of noise to the system: isotropic Gaussian noise and non-linear noise (\autoref{fig:Figure2_new}b).
%
In the case of Isotropic (additive) Gaussian noise, noise was added as:
\begin{equation}
    y_i = x_i + \epsilon_i, \quad \epsilon_i \sim \mathcal{N}(0, \sigma^2)\,,
\end{equation}
resulting in homoscedastic noise with constant conditional variance:
\begin{equation}
    \text{Var}(y_i \mid x_i) = \sigma^2
\end{equation}
%
In non-linear (multiplicative or Heteroscedastic Gaussian) noise, the noise is scaled with the signal magnitude:
\begin{equation}
    y_i = x_i + \sigma \cdot |x_i| \cdot \epsilon_i, \quad \epsilon_i \sim \mathcal{N}(0, 1)\,,
\end{equation}
leading to a signal-dependent variance:
\begin{equation}
    \text{Var}(y_i \mid x_i) = \sigma^2 x_i^2\,.
\end{equation}
This model captures heteroscedasticity often found in real-world systems.

We then evaluated SVHT’s effectiveness in guiding rank selection, both for the case of Isotropic noise (\autoref{fig:Figure2_new}c) and non-linear noise
(\autoref{fig:Figure2_new}d).  
We made the assessment based on comparing the Mean Squared Error (MSE) of DMD-based reconstruction for ranks chosen via SVHT (\autoref{fig:Figure2_new}c-d, left) and ranks chosen via AIC (\autoref{fig:Figure2_new}c-d, right).
In both noise conditions, SVHT tended to select ranks near the "knee point" of the MSE curve—balancing model complexity and reconstruction accuracy. 
We also evaluated how increasing noise levels affect our proposed method’s ability to determine the optimal rank. To do so, we gradually reduced the SNR and examined how the detected rank, comparing SVHT’s estimates with those derived from the MSE curve and from AIC. Across all noise conditions, SVHT consistently selected a rank near the "knee point" (\autoref{fig:Figure2_new}e, darker colors with higher SNR).

Overall, these analyses suggest that SVHT can reliably determine the optimal rank or effective dimensionality of the underlying latent dynamics. 
Notably, SVHT is as good as traditional methods (such as, MSE and AIC/BIC), with more computational efficacy (\autoref{fig:Figure2_new}e).

\input{fig2_new}

\subsection{Assessing novel optimizers for dynamic alignment comparison}
\label{sec:fastDSAvariants}

We first demonstrate that all \fastDsaTermAbbrivPlu\ optimizers we introduce for the alignment problem discussed earlier are superior to previous methods for dynamic similarity analysis \citep{ostrow2023beyond}.
Thus, we compare and characterize the variability across different types of optimizers. 
To conduct this comparison and characterize the differences, we first designed an experiment to quantify speed–accuracy trade-offs under closely matched conditions.
The testbed consisted of generating random square Gaussian matrices with size \(n\in\{2,4,8,16,32,64,128,256\}\). 
For each \(n\), we first generated synthetic symmetric positive definite (SPD) matrices with controlled spectral properties. 
Given a matrix size \(n\) and a minimum eigenvalue threshold \(\lambda_{\min} > 0\), we drew a random matrix \(A \in \mathbb{R}^{n \times n}\) with independent standard normal entries and symmetrized it as
\begin{equation}
    S = \frac{1}{2}(A + A^\top)\,,
\end{equation}
ensuring that \(S\) is real symmetric. We then computed the eigen-decomposition
\begin{equation}
    S = U \, \mathrm{diag}(w) \, U^\top\,,
\end{equation}
where \(U \in \mathbb{R}^{n \times n}\) is orthogonal and \(w = (w_1, \ldots, w_n)\) is the vector of eigenvalues of \(S\). To enforce positive definiteness and a prescribed spectral lower bound, each eigenvalue was replaced by
\begin{equation}
    \tilde{w}_i = |w_i| + \lambda_{\min}\,.
\end{equation}
Finally, the SPD matrix was reconstructed via
\begin{equation}
    A_{\mathrm{SPD}} = U \, \mathrm{diag}(\tilde{w}) \, U^\top\,.
\end{equation}
By construction, \(A_{\mathrm{SPD}}\) is symmetric positive definite and satisfies
\begin{equation}
    \lambda_{\min}(A_{\mathrm{SPD}}) \ge \lambda_{\min}\,.
\end{equation}
This condition is required so that \(A\) is uniformly bounded away from singularity across all matrix sizes.

To obtain pairs of SPD matrices that are related by an orthogonal similarity transform, we proceeded as follows. First, an SPD matrix \(A_1\) was generated using the procedure above. Independently, a random matrix \(M \in \mathbb{R}^{n \times n}\) with standard normal entries was sampled, and its QR decomposition was computed:
\begin{equation}
    M = C R\,,
\end{equation}
where \(C \in \mathbb{R}^{n \times n}\) is orthogonal. The second matrix in the pair was then defined as
\begin{equation}
    A_1 = C A_2 C^\top\,.
\end{equation}
Because orthogonal similarity preserves eigenvalues and positive definiteness, \(A_2\) is also SPD and satisfies
\begin{equation}
    A_2 \sim A_1 \qquad \text{(orthogonal similarity)}\,.
\end{equation}
This construction provides controlled pairs \((A_1, A_2)\) with identical spectra and a known orthogonal transformation \(Q\), which are well-suited for evaluating our optimization algorithms for the alignment problem. 

We assessed the performance of the tested methods by computing four summary measures and showed that all were superior to DSA, with some variability among them. 
We included the following summary measures:
(1) the alignment loss, \(\|A_1 - C A_2 C^\top\|_F\) (Euclidean score), 
(2) an angular discrepancy between \(A_1\) and the aligned \(C A_2 C^\top\) to capture qualitative misalignment, 
(3) the orthogonality residual \(\|C^\top C - I\|_F\), and 
(4) wall–clock runtime.
All hyperparameters, initializations, and stopping criteria were held fixed across methods.
%
%
The runtime analysis indicated that the regularization-based variant and the Landing algorithm were the most computationally efficient across varying matrix sizes, with the regularization variant being marginally faster (\autoref{fig:figure3_new}a). The Riemannian method was faster than the DSA optimizer for smaller matrices ($n \approx 32$) but exhibited a steeper runtime growth, eventually surpassing DSA's computational cost for larger values of $n$.


We evaluated the magnitude of the residual error in Frobenius (\autoref{fig:figure3_new}b) and angular (\autoref{fig:figure3_new}d) matrices, for both error measures. 
The regularization and Riemannian variants exhibit worsening performance as matrix size increases, while the Landing algorithm shows decreasing error with growing \(n\), indicating improved stability in higher dimensions.
To probe accuracy (dis)similarity estimation more directly, we also examined how often each method correctly identifies a (numerically) zero distance in Euclidean (\autoref{fig:figure3_new}c) and angular (\autoref{fig:figure3_new}e) metrics (we counted dissimilarity estimated less than \(10^{-3}\) as zero). 
For this “zero-detection” frequency, both the regularization variant and the Riemannian method degrade rapidly beyond \(n=2\), whereas the Landing algorithm maintains a substantially higher rate of correct zero identifications and clearly outperforms DSA across sizes.
Taken together, our analyses support the Landing-based optimization for solving the alignment problem as the preferred choice in most cases.
It matches or improves upon the alignment accuracy of previous dynamic similarity metrics, consistently outperforms both previous methods \citep{ostrow2023beyond} and the penalty-based and Riemannian variants in wall–clock time, and maintains robust behavior across a wide range of matrix sizes.

\subsection{Assessing efficacy and reliability of \fastDsaTermAbbriv}
\label{sec:fastDSA-effic-assess}

To demonstrate that the family of \fastDsaTermAbbriv\ methods is both efficient and reliable for dynamic similarity analysis, we compare them under diverse scenarios. 
We apply them to two major sets of simulations, where the underlying dynamics are well characterized a priori (and therefore can serve as a ground truth),
and investigate the (in)sensitivity, convergence, computational efficacy, and accuracy of all \fastDsaTermAbbrivPlu\ variants to different forms of change in governing dynamics.

\subsubsection{Invariance under geometric deformations}

\input{fig3_new}
We first investigated the insensitivity of \fastDsaTermAbbrivPlu\ variants to purely geometrical transformations (i.e., those that leave the underlying dynamics intact).
More specifically, we investigate whether \fastDsaTermAbbriv\ maintains invariance under geometric deformations that preserve the attractor topology. 

The application of all \fastDsaTermAbbrivPlu\ methods on geometric deformations that preserve the attractor structure suggests that \fastDsaTermAbbriv, similar to previous methods \citep{ostrow2023beyond},  maintains invariance under these transformations. 
We compared the performance and computational efficiency of \fastDsaTermAbbriv\ with previous methods \citep{ostrow2023beyond} based on simulation of a neural system known to represent head direction in the brain, which implements a ring attractor topology embedded in a network of $n$ neurons \citep{wang2022multiple}. 
We use a conventional network architecture and drive the network using constant input and dynamic noise.
Dynamics of the network result in an activity bump moving around the ring attractor at a roughly constant velocity (\autoref{fig:fig4_new}a, left):
\begin{equation}
\tau \frac{ds}{dt} = -s + W^T \phi(s) + A \pm \gamma b(t) + \xi(x, t)\,,
\end{equation}
where $s$ represents synaptic activations, and $\phi$ is the neural transfer function that converts synaptic activations into firing rates. The nonlinearity was provided by a Rectified Linear Unit (ReLU). 
To ensure robust dynamics for path integration, two rings with slightly offset connectivity profiles were implemented, distinguished by the $\pm \gamma b(t)$ term. $A$ represents the resting baseline input to the neurons, $b$ describes the driving input that the network integrates, $\gamma$ is the coupling term, $\tau$ is the time constant and $\xi$ denotes the noise term of activity. The connectivity matrix $W$ implements a continuous attractor architecture, primarily characterized by local inhibitory interactions, which ensures stable and robust dynamics for path integration.  
Lastly, we randomly sampled (with a uniform distribution) the network size within the range $n \in [100, 250]$ in each trial.
We then applied a continuous geometric deformation that preserved the ring topology while altering the geometry of the neural activation $r$ via a sigmoid transformation:
\begin{equation}
r = \frac{1}{1 + \exp(-\beta s)}\,.
\end{equation}
Here, $\beta$ controls the bump width, with larger $\beta$ values creating narrower bumps, effectively increasing the ring's geometric deformation.

Our results suggest that all members of the \fastDsaTermAbbrivPlu\ family perform equally well as previous methods \citep[DSA and Procrustes analysis][in terms of accuracy]{ostrow2023beyond}, while being computationally more efficient. 
Comparing different variants of \fastDsaTermAbbrivPlu\ and DSA demonstrates that all methods remain invariant across geometric deformations (\autoref{fig:fig4_new}a, middle; red, green, and purple lines). 
In contrast, the Procrustes method rapidly diverges with higher values of \( \beta \) (\autoref{fig:fig4_new}a, middle; yellow). 
%
Furthermore, our results suggest, on average, \fastDsaTermAbbrivPlu\ computations are at least an order of magnitude faster ($\sim 15\times$) than DSA (\autoref{fig:fig4_new}a, right).

The optimization behavior of the \fastDsaTermAbbriv\ family appears to mirror that of DSA, suggesting comparable optimization landscapes and mathematical properties.
To evaluate the convergence behavior of different \fastDsaTermAbbriv\ methods in comparison to DSA, we conducted a series of numerical experiments. We selected two distinct deformation magnitudes, $\beta = 1$ and $\beta = 4$ (from \autoref{fig:fig4_new}a, middle). In both cases, we anticipated low dynamical similarity distances, reflecting minor or no topological changes. 
%
All \fastDsaTermAbbriv\ methods show a decreasing loss (\autoref{fig:fig4_new}b, c left) and lower similarity scores (\autoref{fig:fig4_new}b, c middle),  reflecting insensitivity to geometrical deformation. 
Results indicated that \fastDsaTermAbbriv methods consistently converged to stable similarity scores, slightly slower than DSA.

Considering convergence and iteration time of \fastDsaTermAbbriv\, the overall computational efficiency of \fastDsaTermAbbriv\ still remains superior to previous methods \citep{ostrow2023beyond}.
To more precisely characterize convergence speed, we analyzed when each algorithm's loss curve flattened by collapsing run-by-run data into a single mean-loss trajectory per method. For each trajectory, we computed the discrete gradient as the absolute difference between successive loss values, $|\Delta \text{loss}|$, and used a small threshold (0.005) to define when changes become negligible. The stopping iteration was then defined as the first point where $|\Delta \text{loss}|$ dropped below this threshold, or alternatively, the iteration with the minimum gradient if no threshold was crossed. 
Despite converging slightly later, \fastDsaTermAbbriv\ variants were faster overall, requiring substantially less computational time per iteration compared to DSA (see, \autoref{fig:fig4_new}b-c, right, 
for each iteration time).

\input{fig4_new}

\subsubsection{Ability to detect shifts in governing dynamics}
On the second step of evaluating \fastDsaTermAbbriv\ efficiency and reliability, we investigate the sensitivity of all \fastDsaTermAbbriv\ methods to changes in system dynamics; in particular, parametric transformations that alter the topology of the attractors in the dynamical system. 
To probe the sensitivity of \fastDsaTermAbbriv\  we applied a parametrically \emph{controlled} deformation of the ring topology into a line attractor within the same network model (\autoref{fig:fig4_new}d, left).
We implement it by progressively ablating neurons at one end of a recurrent neural network governed by a cosine connectivity kernel. 
We defined a continuous deformation parameter:
\begin{equation}
\alpha = 1 - \frac{c}{\ell}\,.
\label{eq:alphaVal}
\end{equation}
where $c$ is the number of ablated neurons and $\ell$ is the interaction length scale. 
When $\alpha = 1$, the network forms a perfect ring, and as $\alpha$ decreases, the attractor becomes increasingly linear. 

Our results demonstrate that all suggested variants are equally sensitive to change in dynamics \citep[as good as previous method, DSA][]{ostrow2023beyond}, while being computationally more efficient. 
Both the family of \fastDsaTermAbbriv\ and DSA sharply increased their estimated distance value as $\alpha$ transitioned from less than one (line attractor) to exactly one (ring attractor), confirming their sensitivity to topological changes (\autoref{fig:fig4_new}d, middle; blue, red, green, and purple traces), in contrast to Procrustes analysis (yellow trace).
Furthermore, comparing the computation times suggests that \fastDsaTermAbbriv\ was approximately 15 times faster than DSA in detecting these topological changes (\autoref{fig:fig4_new}d, right).


Comparing optimization behavior of the \fastDsaTermAbbriv\ family with DSA suggests the \fastDsaTermAbbriv\ family has comparable optimization landscapes and is computationally more efficient (even after taking into account the convergence speed).
We assessed the performance of \fastDsaTermAbbriv\ under varying topological transformations by selecting two values of $\alpha$ (\autoref{fig:fig4_new}e,f): $\alpha = 0.2$, corresponding to minor topological changes (\autoref{fig:fig4_new}e), and $\alpha = 0.9$, indicating more substantial topological modifications (\autoref{fig:fig4_new}f). 
As expected, all \fastDsaTermAbbriv\ methods show a decreasing loss (\autoref{fig:fig4_new}e,f left) and lower similarity scores were observed at $\alpha = 0.2$, while higher scores emerged at $\alpha = 0.9$, clearly reflecting sensitivity to the degree of topological transformation (\autoref{fig:fig4_new}e, middle). 
The stopping iterations reveal that \fastDsaTermAbbriv methods generally converge slightly later than DSA for the selected tasks (\autoref{fig:fig4_new}e, f left; comparing the corresponding vertical lines).
%
However, despite the slightly slower convergence of \fastDsaTermAbbriv\ in terms of iteration count, the per-iteration computation of \fastDsaTermAbbriv\ is considerably shorter, resulting in a reduction in overall runtime (approximately 100 times faster than DSA; see, \autoref{fig:fig4_new}e, f right ). 

\input{fig2}

\subsubsection{Sensitivity to fine changes in governing dynamics}
\label{sec:sensitiv-fine-change}
Lastly, we investigate whether \fastDsaTermAbbriv\ is sensitive to fine changes in dynamics and whether this sensitivity is superior to previously suggested alternatives \citep[e.g., Wasserstein-based methods, ][]{kamiya2024koopman, redman2024identifying}. 
A key advantage of \fastDsaTermAbbriv\ compared to alternative routes \citep[in terms of computational efficiency, ][]{redmanconjugacy,redmanneurips, kamiya2024koopman} is its sensitivity to \emph{minor} changes in the dynamics of the system, while maintaining the robustness to noise.
The capacity of \fastDsaTermAbbriv\ to maintain sensitivity, as elaborated below, stems from its multifaceted approach to computing dynamic similarity. 
The \fastDsaTermAbbriv\ family takes into account \emph{both} eigenvalues \emph{and} eigenvectors to compare the governing dynamics; whereas Wasserstein-based approaches \citep{redmanconjugacy,redmanneurips, kamiya2024koopman} focus only on eigenvalue distributions.
Thus, as long as the eigenvectors shaping the flow fields are sufficiently structured and distinct, even in the presence of noise, the governing dynamics can be discriminated.

We first introduce one of the key routes for accelerating the estimation of dynamic similarity \citep[discussed in, ][]{kamiya2024koopman}, which is based on a combination of KernelDMD \citep{Williams_2015, baddoo2022kernel}, and using the Wasserstein distance as the similarity measure (we refer to it by \emph{kwDSA} in the following), which brings more computational efficiency to dynamic similarity analysis. 
We then demonstrate that this alternative approach, kwDSA, is nevertheless less sensitive to changes in dynamics compared to all variants of \fastDsaTermAbbriv\ we introduced.
%
Employing kernel-based methods \citep{bernhard2004, bernhard2001} and KernelDMD \citep{gonzalez2021kernel}, more specifically, using kernel functions (e.g., Radial Basis Function or RBF) to implicitly map data into a high-dimensional feature space (known as reproducing kernel Hilbert space),  
considerably enhances computational efficiency, particularly for large-scale or high-dimensional data. 
This flexibility stems from representing complex nonlinearities without explicitly mapping to the high-dimensional space. By combining KernelDMD with the Wasserstein distance, the similarity between systems can be assessed by comparing the distributions of their spectral components, typically the eigenvalues or singular values of the corresponding Koopman operators \citep{kamiya2024koopman}.  
We calculated KernelDMD as  
\begin{equation}
    A_v = \frac{V^\top K_{YX} U}{N}\,,
\end{equation}
where $V$ and $U$ are the low-rank orthonormal basis matrices (from SVD or eigendecomposition), $K_{YX}$ is the cross-kernel matrix between delayed $Y$ and $X$, and $N$ is the number of time steps or samples.  

The Wasserstein distance provides a principled way to compare probability distributions and has been used to quantify spectral differences between dynamical systems.  
A recent study \citep{redman2024identifying} demonstrated that Wasserstein-based metrics can capture dynamical similarity across nonlinear systems, showing both stability and interpretability when comparing spectral measures derived from Koopman or DMD operators.  
This suggests the Wasserstein distance could be a powerful tool for distribution-based comparisons.
However, as we elaborated below, focusing solely on eigenvalue distributions can obscure important geometric differences encoded in the eigenvectors and the underlying flow fields.
%
Thus, it possibly makes the dynamic similarity estimation less sensitive to differences that lie in the configurations of \emph{eigenvectors} and the associated vector fields. 
Notably, the distribution of eigenvalues might be very similar across systems, even when the corresponding eigenvectors (which encode the flow directions) differ. 
To demonstrate the potential shortcomings of kwDSA, we compared the results of kwDSA with \fastDsaTermAbbriv\ variants. 
 
Our results suggest that kwDSA is less sensitive to the transition in the regime of the system's dynamics as the parameter $\alpha$ approaches one (\autoref{fig:Figure5_noisy}a).
The advantage of \fastDsaTermAbbriv\ over kwDSA is even more pronounced in the presence of noise in the data.
To evaluate robustness to noise, we repeated the ring-to-line attractor transition experiment (presented in \autoref{fig:fig4_new}d) while adding Gaussian white noise to the input data (observation noise). We tested four different signal-to-noise ratios (SNRs): 855, 213, 8.3, and 0.6 (corresponding to progressively increasing levels of noise). 
More specifically, zero-mean white noise was added directly to the neural activity trajectories before applying DMD-based computations. Across all noise levels, \fastDsaTermAbbriv\ consistently detected the topological transition (as $\alpha$ approached 1), while kwDSA was less sensitive (\autoref{fig:Figure5_noisy}a; detection capacity degraded after SNR 8.3, second column).  
These results further highlight \fastDsaTermAbbrivPlu\ robustness to input variability and its advantage in capturing dynamical changes, even under noisy conditions, compared to fast alternatives, such as kwDSA.
%
%
Although kwDSA can serve as a reasonable approximation in certain cases \citep[for instance, where comparing only the eigenvalue spectra is sufficient, and we provided an improved implementation of kwDSA in our \texttt{\fastDsaTermAbbriv}\ package,][]{behradcmclab}, it might not be sufficiently sensitive to fine changes in the system's dynamics; for instance (as discussed in the following) when the dynamics of the systems under comparison only have fine differences in their flow fields.

Lastly, to explicitly demonstrate the mentioned limitation of kwDSA for dynamic similarity estimation (insensitivity to differences in eigenvectors), we also took an analytical approach.
We defined two low-dimensional nonlinear systems (system A and system B) with identical eigenvalues but distinct eigenvectors (\autoref{fig:Figure5_noisy}b). 
Analytical verification of these spectral differences is straightforward due to the simplicity of the system’s dynamics (see section \nameref{sec:twoSystem}, for further details). 
As we discussed earlier, we expect that Wasserstein-based dynamics similarity analysis is less sensitive to differences between such pairs of dynamical systems as they share identical eigenvalues. 
However, we expect \fastDsaTermAbbriv\ to be sensitive, given that it also considers the flow fields.  
We empirically validate our hypothesis through simulating these dynamical systems. 
Given the sensitivity of such systems to the initial condition, we ran the simulations with multiple initial conditions (20 uniformly sampled points near the origin, 80 simulations). 
We then used Multidimensional Scaling \citep[MDS,][]{cox2008multidimensional} to represent the distance (dissimilarities) derived from \fastDsaTermAbbriv\ in a 2-dimensional space.
Our results suggest that \fastDsaTermAbbriv\ variants correctly distinguish systems A and B (\autoref{fig:Figure5_noisy}c-d), whereas kwDSA struggles to differentiate them (\autoref{fig:Figure5_noisy}f). 
Furthermore, \fastDsaTermAbbriv\ (compared to kwDSA) maintains this separability across a wide range of parameters of our simulation. 
Given that the separability of these dynamical systems depends on simulation parameters (e.g., the weight of nonlinear terms, $\epsilon$),
we used Support Vector Machine \citep[SVM, ][]{hearst1998support} with an RBF kernel to separate these two systems across different choices of parameters of our simulations.
Our results demonstrate that \fastDsaTermAbbriv\ provides superior clustering accuracy compared to kwDSA (\autoref{fig:FiguretwoSystemAccuracyStat}).
Thus, these results suggest that \fastDsaTermAbbriv\ gains efficiency yet remains sensitive to fine changes in underlying dynamics, 
while alternative routes, based on Wasserstein-based distance, are less sensitive to changes in dynamics.

\section{Discussion}
In this study, we introduce a novel similarity metric, \fastDsaTermAbbriv, that enables comparing diverse nonlinear dynamical systems at large scales. 
In particular, the computational efficiency and accuracy of \fastDsaTermAbbriv, allow comparing dynamics across neural systems with diverse characteristics (e.g,. architectures, input regimes, and learning curricula) to understand their convergent and divergent computational principles \citep{mante2013context, vyas2020computation, luo2025transitions, pagan2025individual, perich2025neural, chaudhuri2019intrinsic, driscoll2024flexible, jensen2024recurrent, ji2025discovering, genkin2025dynamics, tolmachev2025single, sohn2019bayesian, safavi2024signatures}. 
As it was shown in previous studies, a core method for understanding computation in neural systems is dynamic similarity analyses \citep{eisen2024propofol, ostrow2024delay, guilhot2024dynamical, codol2024brain, bolager2024gradient, huang2024measuring, xu2025nn, huang2025inputdsa} and
\fastDsaTermAbbriv\ allows these crucial analyses to scale up by achieving high computational efficiency, while maintaining the accuracy of previous metrics \citep{eisen2024propofol, ostrow2024delay, guilhot2024dynamical, codol2024brain, huang2024measuring, huang2025inputdsa}. 
To realize \fastDsaTermAbbriv, we incorporate ideas from random matrix theory \citep{plerouRandomMatrixApproach2002,paulRandomMatrixTheory2014,leeRandomMatrixTheory2014,izenmanRandomMatrixTheory,karouiSpectrumEstimationLarge2008}, namely,  our automated rank estimation through SVHT \citep[][]{gavish2014optimal} and three different variants of accelerated optimizers for the alignment of vector fields. 
Furthermore, \fastDsaTermAbbriv\ reduces computational complexity substantially, while maintaining sensitivity to even fine differences in the dynamics and also being insensitive to purely geometrical changes in the activity.
We demonstrate the efficiency, robustness, and sensitivity of all \fastDsaTermAbbriv\ variants through a number of numerical experiments; 
namely,
geometric deformations of a ring attractor network, showing \fastDsaTermAbbriv\ methods remain invariant to topology-preserving transformations while being at least  \(\sim\)150x faster than previous methods \citep{ostrow2023beyond}; 
controlled topological transitions from ring to line attractors, where \fastDsaTermAbbriv\ detected changes as effectively as previous methods \citep{ostrow2023beyond} but at least \(\sim\)15x faster; 
and also sensitivity to fine-scale dynamical changes, where \fastDsaTermAbbriv\ outperformed previously proposed fast alternatives (kwDSA) in distinguishing systems with identical eigenvalues but different eigenvectors, proving superior sensitivity to subtle differences in underlying dynamics. 
We also drew practical guidance on which alignment optimizer to use when deploying \fastDsaTermAbbriv\ at scale.
Taking into account the dimensionality of the data, the regularization-based solver and the Landing solver achieved the lowest runtimes, with the regularization solver marginally faster in some regimes.
However, the Landing solver exhibited the most stable alignment quality as dimensionality increased.
By contrast, both the regularization-based and Riemannian solvers showed degraded reliability beyond small dimensions, and the Riemannian solver additionally exhibited steeper runtime growth that can outweigh its gains at small sizes.
Taken together, these findings support the Landing-based optimizer as the default choice for most applications of \fastDsaTermAbbriv, particularly in higher-dimensional settings and large-sample studies, with the other solvers primarily serving as useful alternatives in small-dimensional regimes or as complementary methods.

The considerable computational efficiency of \fastDsaTermAbbriv\ facilitates its application in extensive, large-scale, and statistical analyses, to ultimately better understand neural computations in neural networks.
For instance, a recent study demonstrates that dynamic similarity analysis can reveal how computations develop in recurrent neural networks by linking evolving dynamical motifs to learned task structure and performance \citep{guilhot2024dynamical}. 
It highlights how architectural features and training conditions influence internal dynamics, providing a powerful lens on neural computation. 
With its improved efficiency, \fastDsaTermAbbriv\ enables such analyses at scale, supporting broader studies across architectures, input regimes, and learning curricula.
%
%

\fastDsaTermAbbrivCap\ can be particularly helpful to understand computations realized in biological neural systems, as analysis of dynamical similarity provides a crucial tool for validating (or violating) computational models of information processing. 
For instance, let us consider the example of a Recurrent Neural Network (RNN) trained on a center-out reaching task, against activity recorded from the primary motor cortex (M1) of a monkey performing the same task \citep{codol2024brain}. 
Traditional comparison methods, which focus on the geometry (or a static picture) of the high-dimensional neural trajectories, often fail because arbitrary scaling or rotation between the biological and artificial state-spaces can lead to a false conclusion of low similarity, even if the underlying computations are identical. 
Dynamic similarity analysis can overcome such caveats by comparing the governing dynamics, exploiting the flow field of the system's dynamics (which, in a way, specifies the instantaneous transformation rule). 
By establishing whether a single linear map can align the dynamics of the RNN with the dynamics of the M1 neuronal populations, dynamic similarity analysis yields a high similarity score if the RNN is implementing the same computational mechanism as the brain, thus moving validation beyond task performance and to a mechanistic understanding of neural computations. 
Notably, the computationally efficient (and accurate) similarity metrics are crucial for modern models and datasets, given the growing body of AI-inspired models in neuroscience \citep[e.g., foundation models, ][]{azabou2024building, zhou2025brain, wang2025foundation, binz2025foundation, wang2025large} and growing complexity and dimensionality of neural data recordings 
\citep{junFullyIntegratedSilicon2017a, juavinettChronicallyImplantedNeuropixels2019, pesaranInvestigatingLargescaleBrain2018, junFullyIntegratedSilicon2017a}.

There are also numerous avenues that can further expand the applicability of the \fastDsaTermAbbriv. 
A key limitation of \fastDsaTermAbbriv\ methods \citep[similar to previous methods][]{ostrow2023beyond} is their reliance on accurately determining embedding parameters, specifically the optimal time-delay and embedding dimension (number of delays), for constructing Hankel matrices ($H$, \autoref{eq:dmdRank}). 
These parameters can influence the estimation of the underlying dynamics.
In our analysis of two low-dimensional nonlinear systems (see section \nameref{sec:sensitiv-fine-change}), 
we employed the Bayesian Information Criterion \citep[BIC,][]{miao2007condition} over a wide range of methods to select suitable embedding parameters (see section \nameref{sec:supp-embed}, \autoref{fig:delayOptimization}, and \stab{Table:OptimizationDelay}). However, while the BIC provides a statistically grounded approach, it is not computationally efficient, which makes the development and validation of reliable, fully automated, and computationally efficient algorithms essential. 
Future studies should focus on devising robust (yet fast) statistical methods for adaptive parameter selection, ensuring faithful dynamic reconstruction across diverse datasets and contexts. 
The same also applies to the optimization procedure, for instance, the regularization weight that we introduced earlier \(\lambda\), as well as learning rate parameters used in Rim- and Land-\fastDsaTermAbbriv.  
Furthermore, although SVHT performs well across the regimes we investigated, it can slightly overestimate the effective rank, and this tendency is most pronounced in very low-noise (high-SNR) settings.
SVHT selects its cutoff from the singular-value spectrum (e.g., $\tau=\omega(\beta)\!\cdot\!{\rm median}$ under unknown noise).
When noise is negligible, the “noise bulk” effectively vanishes and the spectrum’s median can become signal-dominated, causing the threshold to drop.
As a result, small components introduced by delay/Hankel embedding and by mild non-orthogonal geometric deformations can pass the cutoff and be counted as additional modes, inflating the estimated rank.
In principle, a per-dataset sweep on all ranks and selects an “elbow/knee” in an MSE curve (see \autoref{fig:Figure2_new}c-e) could serve as a gold standard for benchmarking rank-selection rules.
However, using such a sweep as a default analysis procedure is often computationally expensive (as it requires DMD-based reconstruction for the entire range of rank candidates), and in many realistic settings, the MSE--rank curve does not exhibit a clearly defined knee point, making such methods an ideal alternative.
These observations motivate future work on rank-selection and preprocessing strategies that remain stable across SNR regimes and across embedding choices, while preserving the practical efficiency that enables large-scale dynamical similarity analyses.
Lastly, for further extending \fastDsaTermAbbriv\ involves integrating Dynamic Mode Decomposition (DMD) variants tailored for noisy or real-world data. Noise-aware methods such as optimized DMD or Bayesian DMD \citep{takeishi2017bayesian} could improve the robustness of \fastDsaTermAbbriv\ when applied to empirical datasets characterized by significant noise levels. Exploring these enhancements would not only bolster the applicability of \fastDsaTermAbbriv\ in neuroscience and artificial intelligence but also extend its utility to a wider range of scientific and engineering fields where dynamic similarity assessment is critical.
To sum, \fastDsaTermAbbriv\ provides a computationally efficient framework for dynamical similarity analysis, is well-suited for large-scale comparisons of neural networks, neuronal circuits, or data with a model, and its design supports future extensions to broader tasks and data modalities.

\section{Methods}
\label{sec:method}

\subsection{Computational foundation of \fastDsaTermAbbriv}
\label{sec:fast-optz-dsa}

Beyond automatic rank detection (discussed earlier), the key component that we introduced to boost the computational efficiency of dynamic similarity analysis is the optimization procedures. 
We develop novel optimizations for aligning the flow field of dynamical systems that allow us to compare neural networks, neuronal circuits, and neural data. 
We discussed the three optimization procedures in the following. 
(1) \fastDsaTermAbbrivCap\ with a regularized optimization (RO-\fastDsaTermAbbriv). 
(2) \fastDsaTermAbbrivCap\ that is optimized based on-manifold Riemannian method (Rim-\fastDsaTermAbbriv).
(3) \fastDsaTermAbbrivCap\ that is optimized based Landing algorithm (Land-\fastDsaTermAbbriv).


\subsubsection{\fastDsaTermAbbrivCap\ with regularized optimization (RO-\fastDsaTermAbbriv)}
\fastDsaTermAbbrivCap\ compares dynamics by aligning the induced operators/flow fields under (near-)orthogonal similarity transforms. In the first variant, we relax the orthogonality constraint on \(C\) during optimization by introducing a penalty term that discourages deviations from orthogonality, and then apply a post–processing step to enforce exact orthogonality at the end. 
Concretely, we minimize the following \emph{regularized} loss:
\begin{equation}
    L(C) = \| A - C B C^\top \|_F^2 + \lambda \| C^\top C - I \|_F^2\,,
    \label{eq:loss_reg}
\end{equation}
where the transformation loss \(\| A - C B C^\top \|_F^2\) measures how well \(C\) maps \(B\) onto \(A\) in Frobenius norm,
\begin{equation}
    \| X \|_F = \sqrt{\sum_{i,j} X_{ij}^2}\,,
\end{equation}
and the regularization term
\begin{equation}
    \lambda \| C^\top C - I \|_F^2
\end{equation}
penalizes departures from orthogonality, controlled by the parameter \(\lambda > 0\). We treat \(\lambda\) as a hyperparameter, though it could be selected by more principled procedures \citep[e.g.,][]{mackevicius2019unsupervised}.
Using gradient descent, we update \(C\) iteratively:
\begin{equation}
    C^{(t+1)} = C^{(t)} - \eta \nabla L\bigl(C^{(t)}\bigr)\,,
    \label{eq:Cupdate}
\end{equation}
where \(\eta\) is the learning rate. The gradient of \(L(C)\) decomposes as
\begin{equation}
    \nabla L(C) = \frac{\partial}{\partial C} \| A - C B C^\top \|_F^2
    + \lambda \frac{\partial}{\partial C} \| C^\top C - I \|_F^2\,,
\end{equation}
with
\begin{align}
    \nabla_C \| A - C B C^\top \|_F^2 &= -4\bigl(A - C B C^\top\bigr) B C 
    \quad \text{(transformation loss gradient)},\\
    \nabla_C \| C^\top C - I \|_F^2 &= 4 \bigl(C C^\top C - C\bigr)
    \quad \text{(orthogonality regularization gradient)}\,.
\end{align}
Thus, the explicit update rule becomes (see section \nameref{Derivative for regularization term} for a detailed derivation of the gradients)
\begin{equation}
    C^{(t+1)} = C^{(t)} - \eta \left[ -4\bigl(A - C B C^\top\bigr) B C
    + 4\lambda \bigl(C C^\top C - C\bigr) \right]\,.
\end{equation}
After optimization, the matrix \(C\) will in general only be approximately orthogonal. We therefore enforce exact orthogonality via a final projection. First, we compute the SVD of \(C\),
\begin{equation}
    C = U \Sigma V^\top\,,
\end{equation}
and set
\begin{equation}
    C_{\text{orthogonal}} = U V^\top\,.
\end{equation}
To ensure that the resulting matrix lies in the special orthogonal group \(SO(n)\), we correct the determinant of \(C_{\text{orthogonal}}\) following \citet{lawrence2019purely}: if \(\det(C_{\text{orthogonal}}) < 0\), we flip the sign of the last column of \(U\) (corresponding to the smallest singular value) and recompute \(C_{\text{orthogonal}} = U V^\top\). This guarantees that \(C_{\text{orthogonal}}\) is a proper rotation (no reflections) and belongs to \(SO(n)\subset O(n)\).
Overall, allowing \(C\) to deviate slightly from strict orthogonality during optimization, smooths the loss landscape and typically speeds up convergence, while the final projection step ensures that the solution exactly satisfies the orthogonality constraint.

\subsubsection{Riemannian \fastDsaTermAbbriv\ (Rim-\fastDsaTermAbbriv)}

We develop the second variant of \fastDsaTermAbbriv\ to avoid soft penalties altogether and to optimize \autoref{eq:lossEuc} directly on the orthogonal manifold. 
Rather than relaxing the constraint and penalizing deviations in Euclidean space, we treat \(O(n)\) itself as the optimization domain: at each iterate \(C\), we project the Euclidean gradient onto the tangent space \(T_C O(n)\) to obtain a Riemannian gradient.  
This implies mapping tangent steps back with a retraction to keep all iterates on (or extremely close to) the family of \(O(n)\), which lead to numerical stability and satisfies the orthogonality constraint without the need for retraction.

The tangent space of \(O(n)\) at a point \(C\) is the linear subspace of \(\mathbb{R}^{n \times n}\) given by
\begin{equation}
    T_C O(n) \;=\; \{\, C S \;|\; S^\top = -S \,\}\,,
\end{equation}
that is, the set of directions obtained by left-multiplying \(C\) with a skew-symmetric matrix.  
Let \(G = \nabla f(C)\) denote the Euclidean gradient of \autoref{eq:lossEuc}. 
Its Riemannian projection onto \(T_C O(n)\) is
\begin{equation}
\label{eq:rgrad}
    \mathrm{grad}\,f(C)
    \;=\;
    \nabla f(C) - C\,\mathrm{sym}\!\bigl(C^\top \nabla f(C)\bigr)
    \;=\;
    \mathrm{skew}\!\bigl(\nabla f(C)\,C^\top\bigr)\,C\,,
\end{equation}
with \(\mathrm{sym}(M)=\tfrac12(M+M^\top)\) and \(\mathrm{skew}(M)=\tfrac12(M-M^\top)\).  
This removes the normal component of the gradient, yielding a descent direction that lies entirely within the tangent space.

At each iteration, we take a tangent step and immediately map it back to the manifold using a retraction:
\begin{equation}
\label{eq:riem-step}
    \Delta_t \;=\; -\eta\,\mathrm{grad}\,f\!\bigl(C^{(t)}\bigr) \;\in\; T_{C^{(t)}}O(n),
    \qquad
    C^{(t+1)} \;=\; \operatorname{Retr}_{C^{(t)}}\!\bigl(\Delta_t\bigr)\,,
\end{equation}
where \(\eta>0\) is the step size.  
This ensures that iterates remain orthogonal up to numerical precision, with no soft penalties and no off-manifold drift.

Several standard retractions can be used to map tangent updates back to the orthogonal manifold:
\begin{align}
\text{Polar:}\quad
&\operatorname{Retr}^{\mathrm{polar}}_C(\Delta)
\;=\;
(C+\Delta)\,\Bigl((C+\Delta)^\top(C+\Delta)\Bigr)^{-\frac12},
\\[3pt]
\text{QR:}\quad
&\operatorname{Retr}^{\mathrm{qr}}_C(\Delta)
\;=\;
\mathrm{qf}(C+\Delta)
\quad\text{(orthogonal factor of a QR with }\mathrm{diag}(R)>0\text{)}, 
\\[3pt]
\text{Cayley (skew update):}\quad
&\operatorname{Retr}^{\mathrm{cayley}}_C(\Delta)
\;=\;
\bigl(I-\tfrac12 S\bigr)^{-1}\!\bigl(I+\tfrac12 S\bigr)\,C,
\qquad
S=\mathrm{skew}\!\bigl(C^\top\Delta\bigr)\,.
\end{align}
Riemannian momentum or Nesterov acceleration can be formulated in the tangent space, and periodic hard reorthogonalization (e.g., invoking the manifold projection every \(K\) steps) can further enhance numerical stability.
Taken together, this construction gives a conceptually distinct variant of \fastDsaTermAbbriv\ in which the orthogonality constraint is not enforced at every iteration by design. In the following sections, we place this strictly manifold-constrained scheme alongside the regularization-based.

\subsubsection{\fastDsaTermAbbrivCap\ with landing-based optimization (Land-\fastDsaTermAbbriv)}

In contrast to the strict on-manifold scheme that was described above, the Landing algorithm \citep{ablin2022fast} follows a manifold-aware optimization strategy that updates \(C\) in directions consistent with the geometry of \(O(n)\), but performs all steps in the ambient Euclidean space. This avoids per-step retractions while keeping iterates close to the orthogonal manifold throughout optimization.
To compute these geometry-aware updates, we again exploit the local structure of \(O(n)\).  
The tangent space of \(O(n)\) at a point \(C\) is
\begin{equation}
    T_C O(n) = \{ C S \; | \; S^\top = -S \}\,,
\end{equation}
i.e., all directions obtained by left-multiplying \(C\) with a skew-symmetric matrix \(S\).  
Let \(G = \nabla f(C)\) denote the Euclidean gradient of the objective in \autoref{eq:lossEuc}.  
A manifold-aware (tangent-like) descent step is obtained by projecting the relative gradient \(G C^\top\) onto the space of skew-symmetric matrices and transporting it back to the ambient space:
\begin{equation}
\label{eq:tangent-like}
    \Psi(C) = \mathrm{skew}(G C^\top) \, C,
    \qquad \mathrm{skew}(M) = \tfrac{1}{2}(M - M^\top)\,.
\end{equation}
The field \(\Psi(C) \in T_C O(n)\) therefore, represents a first-order Riemannian descent direction on \(O(n)\), steering updates that preserve orthogonality to infinitesimal order.
Because our optimization operates in the full Euclidean space (see section~\nameref{Derivative of Landing algorithm}), additionally, a lightweight potential that softly penalizes deviations from orthogonality has been introduced:
\begin{equation}
\label{eq:landing}
    \Lambda_{\mathrm{L}}(C) = \lambda_{\mathrm{L}} (C C^\top - I) C, \qquad \lambda_{\mathrm{L}} > 0.
\end{equation}
This landing term continuously attracts \(C\) back toward the orthogonal manifold by reducing the residual \(R(C) = C C^\top - I\).  
Even though we do not apply a retraction at every step, explicit Euler updates drift off \(O(n)\) only by \(O(\eta^2)\); the field \(-\Lambda_{\mathrm{L}}\) approximates the negative gradient of \(\tfrac{1}{2}\|R(C)\|_F^2\). 
So for sufficiently small \(\eta>0\) the quantity \(\|R(C)\|_F^2\) decreases, and iterates remain near-orthogonal without hard constraints at each iteration.
Combining the tangent-like descent from \autoref{eq:tangent-like} with the landing potential in \autoref{eq:landing} yields the hybrid update rule:
\begin{equation}
\label{eq:hybrid-step}
    C^{(t+1)} = C^{(t)} - \eta \Bigl( \mathrm{skew}(G C^\top) \, C \;+\; \lambda_{\mathrm{L}} (C C^\top - I) C \Bigr),
\end{equation}
which balances movement along orthogonal directions with a soft feasibility correction.  
When exact orthogonality is required after convergence, a single projection—obtaining the orthogonal factor from a QR or polar decomposition—can be applied at negligible computational cost.  
If a strictly feasible rotation is needed, one may further adjust the sign of the last column to ensure \(\det(C)=1\), thus projecting onto \(SO(n)\).

Overall, the Landing-based \fastDsaTermAbbriv\ solver retains the operator-alignment objective but replaces repeated on-manifold retractions with Euclidean-space updates guided by geometry-aware directions, followed by projection onto the orthogonal group when required. This substantially reduces computational overhead while preserving alignment quality and dynamical (dis)similarity estimates.

\subsection{A heuristic optimality for a pair of Hankel embeddings}
\label{sec:supp-minmax}

In our pipeline, we estimate the rank of the Hankel–embedded data for each dataset separately using the optimal SVHT (see sections \nameref{sec:method} and \nameref{sec:supp-SVHT}). Let $H^{(1)}$ and $H^{(2)}$ denote the two Hankel matrices (e.g., from two systems or two views of the same system), with SVHT-selected ranks $r_1$ and $r_2$. This raises a practical question for subsequent linear modeling (DMD/Koopman estimation): should we propagate the \emph{minimum} rank $r_{\min}=\min(r_1,r_2)$ to avoid overfitting, or the \emph{maximum} rank $r_{\max}=\max(r_1,r_2)$ to avoid information loss?

To study this choice and justify our heuristic, we synthesized Lorenz-based data ($\sigma=10$, $\rho=28$, $\beta=8/3$, $dt=0.01$) and added multiplicative (signal–dependent) noise with scale $\sigma=5$ (noise model as in \nameref{sec:supp-SVHT}). We then embedded the noisy series, fit DMD models over a range of candidate ranks $r$, and evaluated reconstruction error on a holdout set via $\mathrm{MSE}(r)$ (see \autoref{fig:minOrmax}). Two robust patterns emerged. First, \emph{underestimating} rank (choosing $r<r_i$) produced a rapid increase in MSE, indicating bias from discarding weak but deterministic components exposed by delay embedding and the nonlinear deformation. Second, modest \emph{overestimation} (choosing $r\gtrsim r_i$) incurred only a mild MSE penalty and often slightly improved performance, presumably by capturing additional coherent structure that sits just above the SVHT cutoff in low-noise regimes.

These analyses suggest that the downside of underestimation was consistently larger than the (small) cost of mild overestimation. 
Thus selecting
\begin{equation}
   r^\star \;=\; r_{\max} \;=\; \max(r_1,r_2)\,, 
\end{equation}
is the safer option, as it avoids sharp MSE degradation while preserving sensitivity to (possibly) weak but meaningful modes. 

This analysis also supports our SVHT-based optimal rank-estimation (see section \nameref{sec:supp-SVHT}). 
\autoref{fig:minOrmax} suggests that slight overestimation of the optimal ranks is less harmful than rank \emph{underestimation}. While modest overestimation may admit weak components (including dynamical components that correspond to noise) into the comparison pipeline, underestimation tends to discard dynamically informative modes and can produce a rapid increase in reconstruction error (MSE), degrading dynamical similarity estimates disproportionately.

\subsection{Two nonlinear dynamical systems with subtle dynamical differences}
\label{sec:twoSystem}
As discussed earlier (section \nameref{sec:sensitiv-fine-change}), to demonstrate the potential shortcomings of Wasserstein-based dynamics similarity analysis, 
we need a simple dynamical system with identical eigenvalues but distinct eigenvectors.  
Thus, we define the two-dimensional dynamical system A based on the following nonlinear Ordinary Differential Equation (ODE):
\begin{align}
\begin{cases}
\dot{x} = -\alpha x + \varepsilon x y,\\[6pt]
\dot{y} = -\beta y - \varepsilon x^2\,.
\end{cases}
\end{align}
We assume small values for parameters $\varepsilon$, and positive constants $\alpha, \beta$. 
Then, system A is a two-dimensional dynamical system with a single equilibrium point at $(0,0)$. The nonlinear terms $\varepsilon xy$ and $-\varepsilon x^2$ introduce mild state-dependent adjustments without causing oscillations or chaos. The linear part of the system is diagonal, meaning that $x$ and $y$ decay independently in the absence of nonlinearities.
Then, the Jacobian of system A at the origin is:
\begin{equation}
    J_A = \begin{pmatrix}
-\alpha & 0 \\[6pt]
0 & -\beta
\end{pmatrix}\,,
\end{equation}
yielding eigenvalues $\lambda_1 = -\alpha$ and $\lambda_2 = -\beta$. The corresponding eigenvectors are the standard basis vectors: $(1,0)^\top$ for $\lambda_1$ and $(0,1)^\top$ for $\lambda_2$.
On the other hand, system B is designed to have the same eigenvalues but different eigenvectors due to its coupling structure:
\begin{align}
\begin{cases}
\dot{x} = -\mu x + \kappa y - \varepsilon x^2,\\[6pt]
\dot{y} = \kappa x - \mu y - \varepsilon y^2\,,
\end{cases}
\end{align}
with parameters selected such that the trace and determinant of the Jacobian match those of $J_A$. We took $\mu = (\alpha+\beta)/2$ and chose $\kappa$ appropriately, ensuring that the eigenvalues of the linear part are $-\alpha$ and $-\beta$ (the same as in system A). The nonlinear terms $-,\varepsilon x^2$ and $-,\varepsilon y^2$ introduce state-dependent damping along each coordinate axis, increasing the dissipation at larger $|x|$ or $|y|$. Like system A, this system has a single equilibrium at the origin and no chaotic behavior. 
Unlike system A, system B includes a symmetric linear coupling between $x$ and $y$ via the $\kappa$ term. This means eigen-directions are no longer aligned with the coordinate axes, but instead lie along linear combinations of $x$ and $y$, altering the flow field of the system despite identical eigenvalues.

\section{Code availability}
The \fastDsaTermAbbriv\ package is available via GitHub at \href{https://github.com/CMC-lab/fastDSA}{https://github.com/CMC-lab/fastDSA} \citep{behradcmclab}.
\section{Acknowledgments}
We would like to thank members of the Computational Machinery of Cognition (CMC) lab, for their discussion and input in the course of development of the \fastDsaTermAbbriv.
%
AB acknowledges the support by the Dresden International Graduate School for Interdisciplinary Life Sciences (DIGS-ILS), and the International Max Planck Research School on Cognitive NeuroImaging (IMPRS CoNI). 
MO was funded by the NSF GRFP.
CB was supported by a Grant from the Volkswagen Foundation Az. 9D175.
SS acknowledges the support through add-on fellowship from the Joachim Herz Foundation.  

\section{Author contribution}

Conceptualization, AB, MO, SS;
Methodology, AB, MO, SS;
Project administration, SS;
Software, AB, MO, MTF;
Formal Analysis, AB, MO, MTF, SS;
Investigation, AB, MO, MTF, SS;
Resources, CB, SS;
Data Curation, AB, MTF;
Writing - Original Draft, AB, SS;
Writing - Review \& Editing: AB, MO, MTF, IF, CB, SS;
Visualization, AB, MO, MTF, SS;
Supervision, IF, SS;
Funding acquisition, CB, SS.


    \bibliography{bibfiles/locallib,bibfiles/tmplib}


\if@endfloat\clearpage\processdelayedfloats\clearpage\fi

\renewcommand\thefigure{\arabic{figure}}
\renewcommand{\figurename}{\bf{Supplementary Figure}}
\renewcommand{\tablename}{Supplementary Table}

\setcounter{figure}{0}  
\setcounter{table}{0}

\pagebreak

\section{Supplementary information}
\label{sec:supp-info}

\subsection{Derivations of optimization update rule for RO-\fastDsaTermAbbriv}
\label{Derivative for regularization term}

We aim to minimize the \autoref{eq:lossEuc} loss function with respect to a transformation matrix $C$, where the first term measures similarity between $A$ and $C B C^\top$, and the second term is an orthogonality regularization that enforces $C \in O(n)$.
For the gradient of transformation loss term, let:
\begin{equation}
    \mathcal{L}_1(C) = \|A - C B C^\top\|_F^2 = \text{Tr}(E^\top E), \quad \text{where } E = A - C B C^\top\,.
\end{equation}
To compute the gradient $\nabla_C \mathcal{L}_1$, using the Frobenius–trace identity $\|X\|_F^2=\mathrm{Tr}(X^\top X)$, we first differentiate:
\begin{equation}
    dE = - (dC) B C^\top - C B (dC)^\top\,.
\end{equation}
Thus, we get:
\begin{equation}
    d\mathcal{L}_1 = 2 \text{Tr}(E^\top dE) = -2 \text{Tr}(E^\top (dC) B C^\top) - 2 \text{Tr}(E^\top C B (dC)^\top)\,.
\end{equation}
Using the transpose/trace rule, 
\[\mathrm{Tr}(X) = \mathrm{Tr}(X^\top)\,,\] \[\mathrm{Tr}(X\,dC^\top) = \mathrm{Tr}(X^\top dC)\,,\]  
and cyclic properties of trace,
\[\mathrm{Tr}(A B C) = \mathrm{Tr}(B C A) = \mathrm{Tr}(C A B)\,,\]
we have:
\begin{equation}
    d\mathcal{L}_1 = -2 \text{Tr}(B C^\top E^\top dC) - 2 \text{Tr}(B^\top C^\top E dC)\,,
\end{equation}
\begin{equation}
    d\mathcal{L}_1 = -2 \text{Tr} \left( \left[ B C^\top E^\top + B^\top C^\top E \right] dC \right)\,.
\end{equation}
Matching with:
\begin{equation}
    d\mathcal{L}_1 = \text{Tr} \left( (\nabla_C \mathcal{L}_1)^\top dC \right)\,.
\end{equation}
Thus, we identify the gradient as:
\begin{equation}
    \nabla_C^{(1)} = (A - C B C^\top) C B^T + (A - C B C^\top)^\top C B\,.
\end{equation}

Then we proceed with the second term, which is defined as:
\begin{equation}
    \mathcal{L}_2(C) = \|C^\top C - I\|_F^2 = \text{Tr}[(C^\top C - I)^2]\,.
\end{equation}
Let us denote:
\begin{equation}
    G = C^\top C - I\,.
\end{equation}
Then, we have:
\begin{equation}
    d\mathcal{L}_2 = 2\ \text{Tr}(G^\top dG) = 2\ \text{Tr}[(C^\top C - I)^\top d(C^\top C)]\,.
\end{equation}
Lastly, using the product rule:
\begin{equation}
    d(C^\top C) = dC^\top C + C^\top dC\,,
\end{equation}
and substituting:
\begin{align}
    d\mathcal{L}_2 &= 2\ \text{Tr}[(C^\top C - I)(dC^\top C + C^\top dC)] \\
                   &= 2\ \text{Tr}[C (C^\top C - I) dC^\top] + 2\ \text{Tr}[(C^\top C - I) C^\top dC] \\
                   &= 2\ \text{Tr}[(C^\top C - I) C^\top dC] + 2\ \text{Tr}[(C^\top C - I)^\top C^\top dC] \\
                   &= 2\ \text{Tr}\left[ \left( (C^\top C - I) C^\top + (C^\top C - I)^\top C^\top \right) dC \right]\,,
\end{align}
we can derive the gradient:
\begin{equation}
    \nabla_C^{(2)} = 2 \left[ C(C^\top C - I) + C(C^\top C - I)^\top \right] = 4(C C^\top C - C)\,.
\end{equation}
Since \( C^\top C - I \) is symmetric, the full gradient is:
\begin{equation}
    \nabla_C \mathcal{L}(C) = \nabla_C^{(1)} + \lambda \nabla_C^{(2)}\,.
\end{equation}
therefore the update rule is \autoref{eq:Cupdate}.

\subsection{Derivations of optimization update rule for Land-\fastDsaTermAbbriv}
\label{Derivative of Landing algorithm}

We minimize the alignment objective
\begin{equation}
    f(C) \;=\; \bigl\|A - C B C^\top\bigr\|_F^2, \qquad C \in O(n)\,,
\end{equation}
and derive the update used by the Landing scheme.  
Let
\begin{equation}
    E \;=\; A - C B C^\top\,,
\end{equation}
so that by the Frobenius–trace identity \(\|X\|_F^2=\mathrm{Tr}(X^\top X)\) we can write
\begin{equation}
    f(C) \;=\; \mathrm{Tr}(E^\top E)\,,
\end{equation}
and its first variation satisfies
\begin{equation}
    df \;=\; 2\,\mathrm{Tr}\!\bigl(E^\top\, dE\bigr)\,.
\end{equation}
Differentiating \(E\) gives
\begin{equation}
    dE \;=\; - (dC)\, B\, C^\top \;-\; C\, B\, (dC)^\top\,,
\end{equation}
hence, using cyclicity of the trace and \(\mathrm{Tr}(X\,dC^\top)=\mathrm{Tr}(X^\top dC)\), we obtain
\begin{equation}
    df \;=\; -2\,\mathrm{Tr}\!\bigl(B\,C^\top E^\top\, dC\bigr)\;-\;2\,\mathrm{Tr}\!\bigl(B^\top C^\top E\, dC\bigr)\,,
\end{equation}
which implies, by matching \(df=\mathrm{Tr}\!\bigl((\nabla f(C))^\top dC\bigr)\), the Euclidean gradient in full generality:
\begin{equation}
    \nabla f(C) \;=\; -2\Bigl(E\, C\, B^\top \;+\; E^\top\, C\, B\Bigr)\,.
\end{equation}
When \(A\) and \(B\) are symmetric, this simplifies to
\begin{equation}
    \nabla f(C) \;=\; -4\,\bigl(A - C B C^\top\bigr)\, B\, C\,,
\end{equation}
but note that the Landing formulation does not rely on this assumption.
To move in directions compatible with the geometry of \(O(n)\), we use the relative skew projection of the Euclidean gradient.  
Defining \(\mathrm{sym}(M)=\tfrac12(M+M^\top)\) and \(\mathrm{skew}(M)=\tfrac12(M-M^\top)\), the Riemannian (projected) gradient on \(O(n)\) can be written as
\begin{equation}
    \mathrm{grad}\,f(C) \;=\; \nabla f(C) \;-\; C\,\mathrm{sym}\!\bigl(C^\top \nabla f(C)\bigr)\,,
\end{equation}
which is equivalently expressed in “relative” form as
\begin{equation}
    \mathrm{grad}\,f(C) \;=\; \mathrm{skew}\!\bigl(\nabla f(C)\,C^\top\bigr)\,C\,.
\end{equation}
For clarity in the non-symmetric case, let \(G=\nabla f(C)=-2(E C B^\top + E^\top C B)\). Then
\begin{equation}
    G\,C^\top \;=\; -2\Bigl(E\, C\, B^\top C^\top \;+\; E^\top\, C\, B\, C^\top\Bigr)\,,\qquad
    (G\,C^\top)^\top \;=\; -2\Bigl(C\, B\, C^\top\, E^\top \;+\; C\, B^\top\, C^\top\, E\Bigr)\,,
\end{equation}
and therefore
\begin{equation}
    \mathrm{skew}(G\,C^\top)
    \;=\; C\, B\, C^\top\, E^\top \;+\; C\, B^\top\, C^\top\, E \;-\; E\, C\, B^\top C^\top \;-\; E^\top\, C\, B\, C^\top\,,
\end{equation}
so the tangent component becomes
\begin{equation}
    \mathrm{skew}(G\,C^\top)\,C
    \;=\; \Bigl(C\, B\, C^\top\, E^\top \;+\; C\, B^\top\, C^\top\, E \;-\; E\, C\, B^\top C^\top \;-\; E^\top\, C\, B\, C^\top\Bigr)\,C\,.
\end{equation}
The Landing vector field augments this tangent descent with a light attraction toward orthogonality.  
Let the orthogonality residual be \(R(C)=C C^\top - I\) and define \(h(C)=\tfrac12\|R(C)\|_F^2\).  
A short calculation yields
\begin{equation}
    \nabla h(C) \;=\; 2\,(C C^\top - I)\,C\,,
\end{equation}
so the landing term
\begin{equation}
    \Lambda_{\mathrm L}(C) \;=\; \lambda_{\mathrm L}\,(C C^\top - I)\,C\,, \qquad \lambda_{\mathrm L}>0\,,
\end{equation}
is proportional to \(\nabla h(C)\) and therefore decreases \(h(C)\) under explicit steps \(C \leftarrow C - \eta\,\Lambda_{\mathrm L}(C)\) for sufficiently small \(\eta>0\).  
Combining the tangent-like direction with the landing pull gives the complete field
\begin{equation}
    \Lambda(C) \;=\; \mathrm{skew}\!\bigl(\nabla f(C)\,C^\top\bigr)\,C \;+\; \lambda_{\mathrm L}\,(C C^\top - I)\,C\,,
\end{equation}
and the explicit Landing update reads
\begin{equation}
    C^{(t+1)} \;=\; C^{(t)} \;-\; \eta\,\Lambda\!\bigl(C^{(t)}\bigr)\,, \qquad \eta>0\,.
\end{equation}
Because no retraction or projection is enforced at each iteration, even when the optimization begins on \(O(n)\), a single explicit Euler step incurs a small drift of order \(\mathcal{O}(\eta^2)\). Thus, the optimization proceeds in the full Euclidean space, with its direction merely biased by the manifold geometry.  
The landing term \(\lambda_{\mathrm L}(C C^\top - I)C\) acts as a soft potential that continuously reduces \(\|C C^\top - I\|_F\) for small steps, counteracting this drift and keeping iterates near the manifold without enforcing feasibility at every step.  
If exact orthogonality is required after convergence, a single projection—such as extracting the orthogonal factor from a QR or polar decomposition—can be performed at negligible computational cost.

In settings where \(C\) is parameterized as a product of factors, \(C_{\mathrm{tot}}=C_1 C_2 \cdots C_m\), it is convenient to express gradients with respect to each factor.  
Let \(L_i=C_1\cdots C_{i-1}\) and \(R_i=C_{i+1}\cdots C_m\) so that \(C_{\mathrm{tot}}=L_i\,C_i\,R_i\).  
A variation \(dC_i\) induces \(dC=L_i\,dC_i\,R_i\), and from \(df=\mathrm{Tr}((\nabla f(C_{\mathrm{tot}}))^\top dC)\) we obtain
\begin{equation}
    G_i \;\equiv\; \nabla_{C_i} f \;=\; L_i^\top\,\nabla f(C_{\mathrm{tot}})\,R_i^\top\,,
\end{equation}
with the corresponding tangent-like direction and landing pull
\begin{equation}
    \Psi_i(C_i) \;=\; \mathrm{skew}\!\bigl(G_i\,C_i^\top\bigr)\,C_i\,, \qquad
    \Lambda_{\mathrm L,i}(C_i) \;=\; \lambda_{\mathrm L}\,(C_i C_i^\top - I)\,C_i\,,
\end{equation}
leading to the per-factor explicit step \(C_i \leftarrow C_i - \eta\bigl(\Psi_i(C_i)+\Lambda_{\mathrm L,i}(C_i)\bigr)\).  
Working via the residual \(E=A-C_{\mathrm{tot}} B C_{\mathrm{tot}}^\top\), one may expand
\begin{equation}
    G_i \;=\; -2\Bigl(L_i^\top\,E\,C_{\mathrm{tot}}\,B^\top\,R_i^\top \;+\; L_i^\top\,E^\top\,C_{\mathrm{tot}}\,B\,R_i^\top\Bigr)\,,
\end{equation}
and substitute into the expression for \(\Psi_i(C_i)\).  
In the symmetric case \(A^\top=A\) and \(B^\top=B\), this simplifies to
\begin{equation}
    G_i \;=\; -4\,L_i^\top\,(A - C_{\mathrm{tot}} B C_{\mathrm{tot}}^\top)\,C_{\mathrm{tot}}\,B\,R_i^\top\,,
\end{equation}
which yields the corresponding simplified per-factor updates.

\subsection{Optimizing delay–embedding parameters}
\label{sec:supp-embed}

Choosing the embedding parameters (the delay interval $\tau$ and the number of delays $\mu$) is a crucial step for identifying the underlying dynamics; however, it is a challenging parameter to tune. In principle, we can use well-established statistical criteria, such as BIC, and perform a grid search over $(\tau,\mu)$ to select the minimum, yielding embeddings that are dependable for downstream analysis. However, exhaustive grids over wide ranges of $\tau$ and $\mu$ are computationally expensive and often impractical. We evaluated several methods that have been suggested to provide reliable parameter selection, including AMI+FNN \citep{wallot2018calculation}, Cao's method \citep{cao1997practical}, Joint Delay–Dimension Optimization  \citep[C--C, ][]{xu2022joint}, and the Unified Embedding Algorithm \citep{kramer2021unified}.

We compared these selectors on a controlled testbed: synthetic Lorenz data ($\sigma=10$, $\rho=28$, $\beta=8/3$, $dt=0.01$), resampled to datasets of size $N\in\{1000,2000,4000,8000,12000\}$, and evaluated by downstream DMD reconstruction on a holdout set. For each method, we recorded wall time, peak memory, the chosen $(\tau,\mu)$, and reconstruction accuracy,
\begin{equation}
    \mathrm{NRMSE} \;=\; \frac{\lVert \widehat{X}-X\rVert_F}{\lVert X\rVert_F}\,.
\end{equation}
Scalability was summarized by fitting log--log slopes of time versus $N$ (see \autoref{fig:delayOptimization} and the accompanying complexity  \stab{Table:OptimizationDelay}).

Across our experiments, BIC (grid over $(\tau,\mu)$) delivered the lowest holdout NRMSE but at higher computational cost and with $\mu$ frequently pushed to the grid ceiling (e.g., at $N{=}1000$, BIC chose $(\tau{=}2,\mu{=}99)$ with NRMSE $\approx 1.5\times 10^{-5}$ in $\sim$6.3\,s; scaling $\sim N^{0.80}$). A projection+ACF heuristic often matched near-BIC accuracy at a fraction of the time, making it a practical default when exhaustive grids are infeasible (e.g., at $N{=}1000$ on one channel it chose $(120,8)$ with NRMSE $\sim 1.9\times 10^{-3}$ in $\sim$0.25\,s), though its selected lags sometimes sat at the scan boundary and performance varied across channels. While AMI+FNN and Cao’s method were consistently fast and stable (sub-second at $N{=}1000$; $\sim N^{0.7\text{--}0.9}$ scaling), they traded prediction fidelity for speed (e.g., NRMSE $\sim 0.18$--$0.35$ at $N{=}1000$). The C--C criterion prioritized topological embedding over forecasting and exhibited the steepest scaling (about $N^{1.5}$), with much longer runtimes (e.g., $\sim$10\,s at $N{=}1000$; $\sim$400\,s at $N{=}12000$) and weaker DMD reconstruction (NRMSE $\sim 0.44$--$0.73$). These outcomes reflect different objectives (forecast accuracy vs.\ speed vs.\ topology) and substantial sensitivity to data channel and sample size.

Taken together, these analyses suggest clear speed--accuracy trade-offs and non-uniform behavior across variables and $N$. At present, none of the procedures provides a universally reliable and scalable choice of $(\tau,\mu)$ across tasks, channels, and sample sizes. A principled, deformation-stable, and forecast-aware embedding selection strategy remains an open problem and warrants deeper investigation in future work.



\clearpage
\section{Supplementary tables}
\input{Table_delayOptimization}

\clearpage
\section{Supplementary figures}
\input{Supp_AlgorithmStructures}
\input{fig_maxOrmin}

\input{fig_Landscape_transformation}
\input{fig_supp_kwDSA}
\input{FigSupp_twoSystem_stat}
\input{fig_delayOptimization}

\end{document}

%% file: config/paperMetaData.tex
\newcommand{\paperTitle}{Fast dynamical similarity analysis}
\newcommand{\shortPaperTitle}{Fast dynamic similarity analysis}

\newcommand{\authorone}{Arman Behrad}
\newcommand{\authortwo}{Mitchell Ostrow}
\newcommand{\authorthree}{Mohammad Taha Fakharian}
\newcommand{\authorfour}{Ila Fiete}
\newcommand{\authorfive}{Christian Beste}
\newcommand{\authorsix}{Shervin Safavi}

\newcommand{\afillA}{Computational Neuroscience, Department of Child and Adolescent Psychiatry, Faculty of Medicine, TU Dresden, Dresden, Germany}
\newcommand{\afillB}{Department of Brain and Cognitive Sciences, Massachusetts Institute of Technology, Massachusetts, USA}
\newcommand{\afillC}{Neural Computation Unit, Okinawa Institute of Science and Technology Graduate University, Okinawa, Japan}
\newcommand{\afillD}{K. Lisa Yang Integrative Computational Neuroscience (ICoN), Massachusetts Institute of Technology, Massachusetts, USA}
\newcommand{\afillE}{Cognitive Neurosphysiology, Department of Child and Adolescent Psychiatry, Faculty of Medicine, TU Dresden, Dresden, Germany}
\newcommand{\afillF}{Department of Computational Neuroscience, Max Planck Institute for Biological Cybernetics, T\"ubingen, Germany}
\newcommand{\afillG}{International Max Planck Research School for Cognitive NeuroImaging (IMPRS CoNI), Leipzig, Germany}

\newcommand{\cauthorEmail}{research@armanbehrad.org, research@shervinsafavi.org}
\newcommand{\cauthorInitials}{AB, SS}

\newcommand{\leadAuthor}{Behrad}


%% file: config/journalConfigs/laPreprint.tex
\documentclass[9pt,biorxiv]{lapreprint}

\usepackage[version=4]{mhchem} 
\usepackage{siunitx}    
\usepackage{pdflscape}  
\usepackage{rotating}   
\usepackage{textgreek}  
\usepackage{gensymb}    
\usepackage[misc]{ifsym} 
\usepackage{orcidlink}  
\usepackage{listings}   
\usepackage{colortbl}   
\usepackage{tabularx}   
\usepackage{longtable}  
\usepackage{subcaption}
\usepackage{multirow}
\usepackage{snotez}     
\usepackage{csquotes}   

\DeclareSIUnit\Molar{M}

\title{\paperTitle}

\author[ \orcidlink{0000-0002-5185-5506} 1,2 \Letter]{\authorone}
\author[ \orcidlink{0000-0002-9732-548X} 3]{\authortwo}
\author[ \orcidlink{0009-0009-3494-3641} 4]{\authorthree}
\author[ \orcidlink{0000-0003-4738-2539} 3,5]{\authorfour}
\author[ \orcidlink{0000-0002-2989-9561} 6]{\authorfive}
\author[ \orcidlink{0000-0002-2868-530X} 1,7 \Letter]{\authorsix}

\affil[1]{\afillA}
\affil[2]{\afillG}
\affil[3]{\afillB}
\affil[4]{\afillC}
\affil[5]{\afillD}
\affil[6]{\afillE}
\affil[7]{\afillF}
\correspondence{\cauthorEmail}{\cauthorInitials}

\leadauthor{\leadAuthor}
\shorttitle{\shortPaperTitle}


%% file: config/journalConfigs/laPreprint_customization.tex
\usepackage{hyperref}
\usepackage{cleveref}

\usepackage{newfloat}

\newcounter{stable}
\newcounter{tempTableCounter}

\crefname{table}{Table}{Tables}
\Crefname{table}{Table}{Tables}
\crefname{stable}{Supp. Table}{Supp. Tables}
\Crefname{stable}{Supp. Table}{Supp. Tables}
\crefname{slongtable}{Supp. Table}{Supp. Tables}
\Crefname{slongtable}{Supp. Table}{Supp. Tables}
\crefname{longtable}{Table}{Tables}
\Crefname{longtable}{Table}{Tables}

\DeclareFloatingEnvironment[fileext=los,name=Supplementary Figure]{sfigure}

\newenvironment{slongtable}[1][]{%
  \setcounter{tempTableCounter}{\value{table}}%
  \setcounter{table}{\value{stable}}%
  \renewcommand{\tablename}{Supplementary Table}%
  \renewcommand{\thetable}{\arabic{table}}%
  \longtable[#1]%
}{%
  \setcounter{stable}{\value{table}}%
  \setcounter{table}{\value{tempTableCounter}}%
  \endlongtable
}

\newcommand{\stab}[1]{%
  \hyperref[#1]{Supp. Table~\ref*{#1}}%
}

\AtBeginDocument{%
}

\crefname{equation}{Eq.}{Eq.}
\crefname{figure}{Fig.}{Fig.}
\Crefname{equation}{Eq.}{Eqs.}
\Crefname{figure}{Fig.}{Figs.}

\providecommand{\Crefname}[3]{}%
\crefname{sfigure}{Supp. Fig.}{Supp. Figs.}
\Crefname{sfigure}{Supp. Fig.}{Supp. Figs.}
\crefname{figure}{Figure}{Figure}

%% file: config/commonConfigStuff.tex
\usepackage{xfrac}

\usepackage{xspace}

\usepackage[textsize = tiny,backgroundcolor=blue!8!white]{todonotes}
\newcommand{\cut}[1]{}

\newcommand{\shsc}[1]{}
\newcommand{\shsd}[1]{}
\newcommand{\shstd}[1]{}

\newcommand{\oltextd}[1]{} 



%% file: config/bibConfigs.tex
\usepackage[numbers,compress,sort,square,comma]{natbib}

%% file: config/journalConfigs/laPreprint_opening.tex
\maketitle


%% file: fig0.tex
\vspace{0.0cm}
\begin{figure}[h!]
\begin{center}
    \hspace{-1.3cm}
    \includegraphics[width=18cm, trim={0cm 0cm 0cm 0cm}]{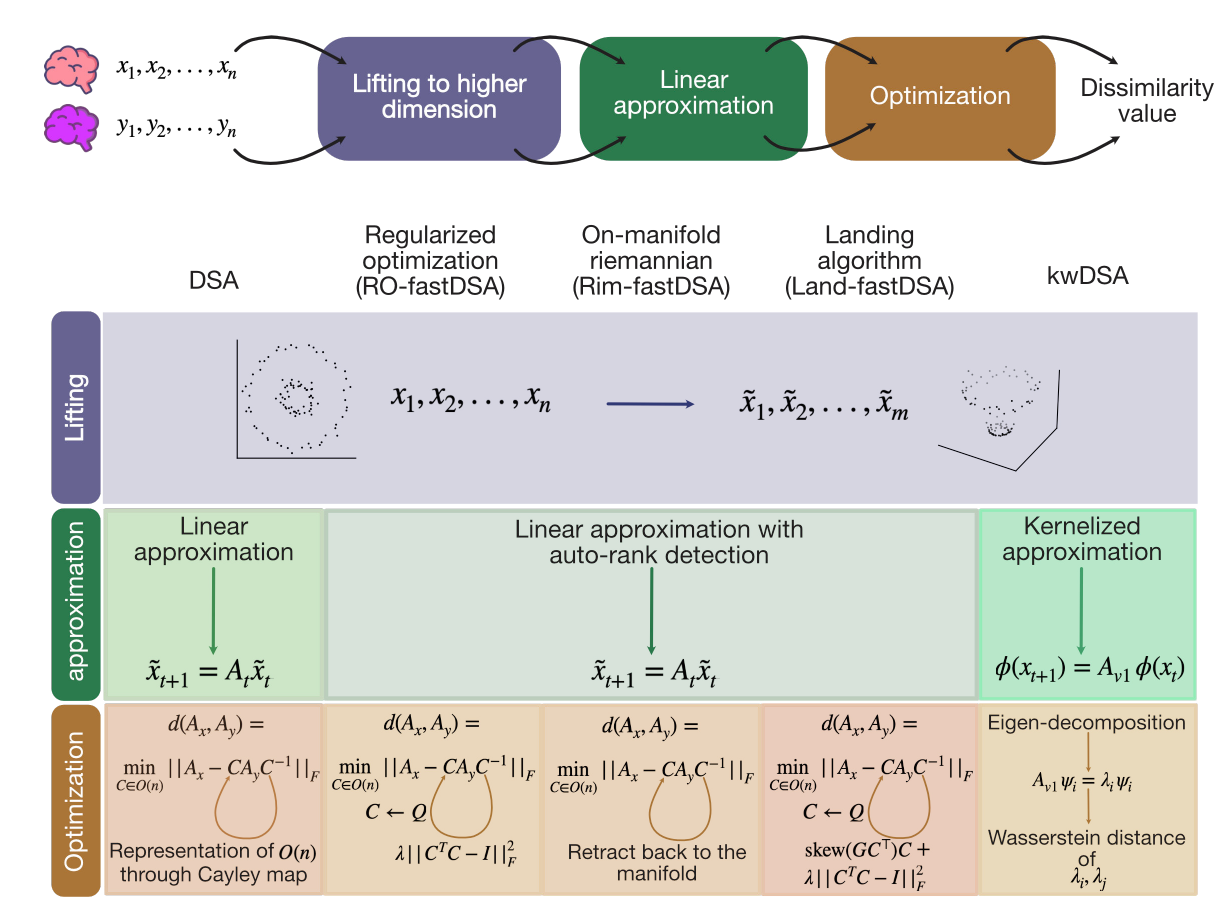}
\end{center}
\vspace{-0.0cm}
\caption{
\textbf{Schematic overview of methods for estimating dynamic (dis)similarity: DSA, family of fastDSA methods, and kwDSA:}
(Top) Depict the overall pipeline common to all methods. 
In particular, the similarity computation is decomposed into three consecutive steps: 
lifting to a higher-dimensional representation (purple), \( (x_1, x_2, \ldots, x_n) \;\longmapsto\; (\tilde{x}_1, \tilde{x}_2, \ldots, \tilde{x}_m)\), where \(m > n\),
linear approximation of the dynamics based on the Koopman theory (green), and optimization of the similarity transform (brown). 
(Bottom) The chart summarizes how each method (DSA, the three fastDSA variants, and kwDSA) instantiates these three stages.
In summary, all three fastDSA variants employ automatic rank detection in the linear approximation step. 
For kwDSA there is an implicit lifting to higher dimension through a kernel function \citep[see section \nameref{sec:sensitiv-fine-change} and ][for more details]{bernhard2004, bernhard2001}. 
Furthermore, for kwDSA, the automatic rank selection can also be applied. 
At the optimization stage, however, the methods differ substantially, including the three fastDSA variants, which leads to distinct computational and accuracy profiles across algorithms (see section \nameref{sec:fastDSA-effic-assess} for more details).
}
\label{fig:fDSAshematic} 
\end{figure}

%% file: fig2_new.tex
\vspace{0.0cm}
\begin{figure}[h!]
\begin{center}
    \hspace{-1.3cm}
    \includegraphics[width=18cm, trim={0cm 0cm 0cm 0cm}]{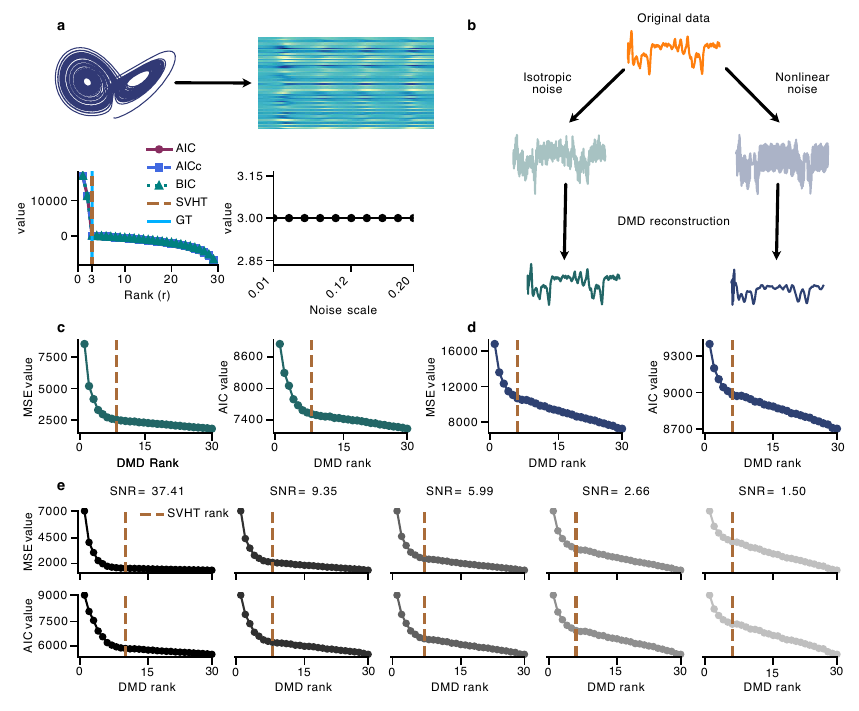}
\end{center}
\vspace{-0.0cm}
\caption{
\textbf{Demonstration of automatic rank reduction with SVHT:}
\textbf{(a)} 
(top-left) Synthetic Lorenz attractor. 
(top-right) Depiction of the projected Lorenz attractor timeseries into a higher-dimensional (128-dimensional) observation space \citep[similar to,][]{kapoor2024latent}. 
X-axis of the heatmap indicates time steps, and the y-axis indicates the dimensions.
(bottom-left) Comparison of optimal rank detection by SVHT (brown broken line) against traditional model selection criteria (AIC, red; AICc, purple; BIC, green). Due to the overlap, they are not all fully visible. 
SVHT accurately identifies the rank of underlying latent dynamics (rank 3) that precisely overlap with the ground truth (GT, vertical continuous blue line).  
AIC, AICc (corrected AIC for small samples), and BIC after the knee point on $r = 3$, exhibit a monotonic decrease in their values with increasing rank $r$. 
(bottom-right) Stability of SVHT-detected optimal rank under varying levels of additive noise. 
\textbf{(b)} 
Schematics of the process of injecting noise to synthetic Lorenz attractor data with isotropic noise (left) or non-linear noise (right), both were generated with variance $\sigma = 5$, and (bottom) DMD reconstructed data using the optimal rank detected by SVHT. 
\textbf{(c-d)} 
Mean Squared Error (MSE), on left, and AIC values, on right, for DMD-based reconstruction across varying levels of (c) isotropic and (d) nonlinear noise. 
\textbf{(e)} 
(top) MSE and (bottom) AIC of DMD-based reconstruction under nonlinear noise, and the SVHT-selected optimal rank (brown broken vertical line) for different values of SNR.
Darker colors correspond to higher SNR values (lower noise levels), and brighter colors to lower SNR values (higher noise levels). SNR values are noted on top of individual columns. 
}
\label{fig:Figure2_new} 
\end{figure}

%% file: fig3_new.tex

\begin{wrapfigure}{r}{.65\textwidth}
  \vspace{-21pt}
  \begin{center}
    \includegraphics[width=12cm, trim={0cm 0cm 0cm 0cm}]{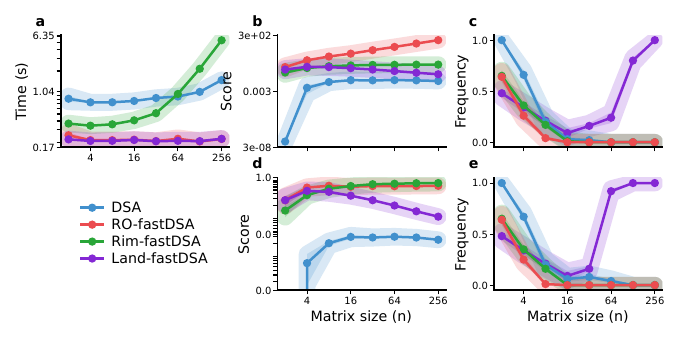}
  \end{center}
  \vspace{-17pt}
\caption{
\textbf{Speed–accuracy trade-offs between DSA and the fastDSA variants under matched experimental conditions.}
\textbf{(a)} Runtime (in seconds) as a function of matrix size for DSA (blue) and the three fastDSA variants (red, RO-fastDSA; green, Rim-fastDSA; purple, Land-fastDSA).    
\textbf{(b)} Frobenius alignment error across matrix sizes for all methods.
\textbf{(c)} Frequency with which each method reports a (numerically) zero Euclidean alignment score (dissimilarity \(< 10^{-3}\)) as a function of matrix size, estimated over 10 independent runs per method and size. 
\textbf{(d)} Angular alignment error across matrix sizes for all methods.  
\textbf{(e)} Frequency with which each method reports an angular score of zero (dissimilarity \(< 10^{-3}\)), indicating almost perfect angular alignment. 
All were estimated over 10 independent runs per method and size.  
In all line plots, shaded regions indicate \(\pm 1\) standard error of the mean across runs.
}
  \vspace{-19pt}
\label{fig:figure3_new} 
\end{wrapfigure}

%% file: fig4_new.tex
\begin{figure}[htbp]
\begin{center}
    \hspace{-1.3cm}\includegraphics[width=18cm, trim={0cm 0cm 0cm 0cm}]{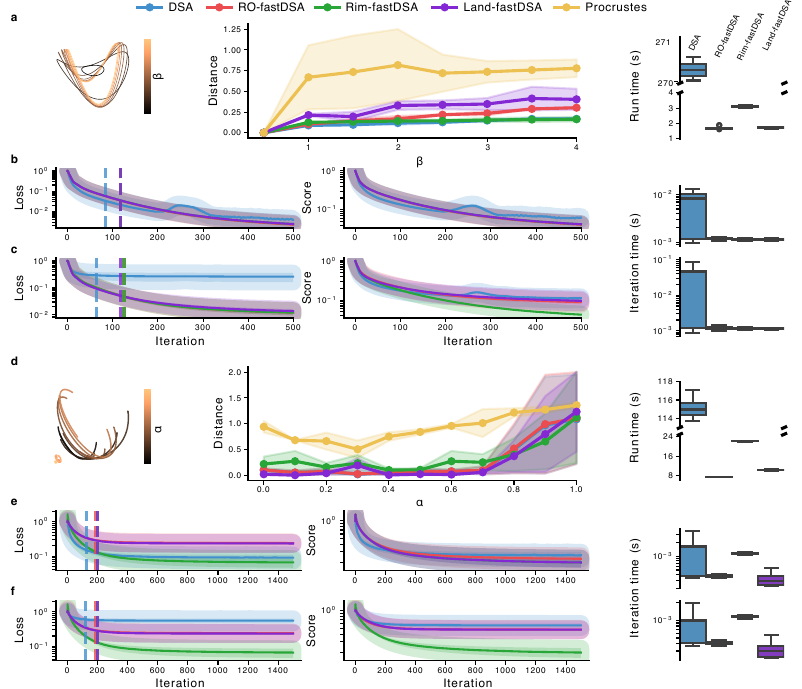}
\end{center}
\vspace{-0.4cm}
\caption{
\textbf{Comparing dynamic similarity estimation and computational efficiency of DSA, fastDSA, and Procrustes analysis.}
Across all panels, blue corresponds to DSA; red, RO-fastDSA; green, Rim-fastDSA; purple, Land-fastDSA.
\textbf{(a)} 
(left) Geometric deformation experiment: Trajectories along the ring are shown in the top three principal components of the network, with intensity of the color indicating the value of $\beta$ (matching the range of the x-axis in the middle figure). Increasing $\beta$ causes the ring to deviate from a planar structure and become increasingly warped in three-dimensional space.
(middle) Dynamical similarity (distance) as a function of \(\beta\), and 
(right) runtime comparison of DSA, the three fastDSA variants. Dynamical similarity remains essentially unchanged under geometric deformation for DSA and all fastDSA variants, whereas Procrustes does not preserve this invariance.
\textbf{(b–c)} 
Normalized optimization trajectories for the geometric case with (b) \(\beta = 1\) and (c) \(\beta = 4\). 
Left: loss over 500 iterations for DSA and the three fastDSA variants. 
Center: difference scores between DSA and each fastDSA variant across iterations. 
Right: Iteration time comparison of DSA, the three fastDSA variants.  
For \(\beta = 1\), the stopping iterations were: DSA 84, regularization 117, Riemannian 117, and Landing 117. 
For \(\beta = 4\), 
the stopping iterations (vertical lines) were: DSA 65, RO-fastDSA 125, Rim-fastDSA 125, and Land-fastDSA 117. 
Corresponding loss and scores without normalization are shown in \autoref{fig:FigureLandscapetransformation}a, and b.
\textbf{(d)} 
(left) Topological transformation experiment: Trajectories along the ring are visualized in the top three principal components of the network activity, where color intensity encodes the value of $\alpha$ (matching the range of the x-axis in the middle figure). 
Increasing $\alpha$ gradually collapses the ring structure into a line, reflecting a topological transformation. 
(middle) Dynamical similarity as a function of \(\alpha\), 
(right) and runtime comparison across DSA and the family of fastDSA variants. 
Sensitivity to the topological change (increase in distance around \(\alpha \approx 0.7\)) is preserved by DSA and all fastDSA variants but not by Procrustes.
\textbf{(e–f)} 
Similar to b-c, optimization trajectories for the topological case with (e) \(\alpha = 0.2\) 
 and (f) \(\alpha = 0.9\). 
Left: loss over iterations; 
center: difference scores between DSA, each fastDSA variant, and 
Right: Iteration time comparison of DSA, the three fastDSA variants. 
For \(\alpha = 0.2\), the stopping iterations were: DSA 129, RO-fastDSA 188, Rim-fastDSA 199, and Land-fastDSA 198. For \(\alpha = 0.9\), the stopping iterations were: DSA 118, RO-fastDSA 188, Rim-fastDSA 199, and Land-fastDSA 198. 
Corresponding loss and scores without normalization are also shown in \autoref{fig:FigureLandscapetransformation}c, and d.
In all line plots, shaded regions indicate \(\pm 1\) standard error of the mean across runs. 
In the box-plots from left to right (also noted on top), DSA, RO-fastDSA, Rim-fastDSA, and Land-fastDSA methods.
}
\label{fig:fig4_new} 
\end{figure}

%% file: fig2.tex
\vspace{0.0cm}
\begin{figure}[htbp]
\begin{center}
    \hspace{-1.3cm}
    \includegraphics[width=18cm, trim={0cm 0cm 0cm 0cm}]{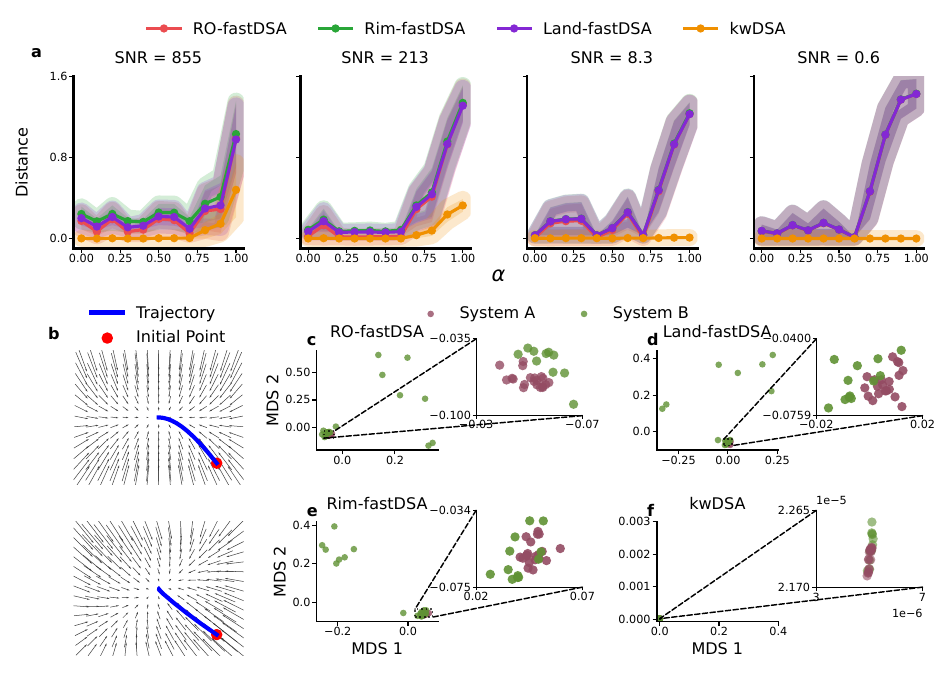}
\end{center}
\vspace{-0.0cm}
\caption{
\textbf{Sensitivity of fastDSA for detecting fine-grained changes in dynamics in the presence of noise:}
\textbf{(a)}
Similarity (distance) as a function \(\alpha\) estimated by the family of fastDSA methods (red, RO-fastDSA; green, Rim-fastDSA; purple, Land-fastDSA), and kwDSA (orange), for four levels of additive Gaussian noise (from left to right, SNR = 855, 213, 8.3, and 0.6). 
All fastDSA methods show greater sensitivity compared to kwDSA in detecting the transition from a line (\(\alpha = 0\)) to a ring attractor (\(\alpha = 1\)).
kwDSA's performance degrades as noise increases (from left to right, higher to lower SNR levels). 
The ranks for each SNR level, detected using SVHT, and they are 156, 174, 68, and 29, respectively. 
For a closer look at the kwDSA results on a logarithmic scale, see \autoref{fig:FigurekwDSA_log}.
Shaded regions in the line plots indicate ±1 standard error of the mean across 10 runs. 
\textbf{(b)}
Flow fields of two nonlinear systems with identical eigenvalues but distinct eigenvectors (see section \nameref{sec:twoSystem} for more details): system A (top) and system B (bottom).
\textbf{(c-f)} 
MDS representation of distances resulted from the (c-e) family fastDSA methods and (f) kwDSA applied on the dynamical systems described in b. 
All fastDSA methods can accurately distinguish the two dynamic systems with fine changes; 
however, kwDSA struggles to separate these systems (see also \autoref{fig:FigurekwDSA_log}).
}
\label{fig:Figure5_noisy} 
\end{figure}

%% file: Table_delayOptimization.tex
\begin{slongtable}{llrrrlr}
\toprule
                  method & channel &   N &   time\_sec &   mem\_MB &  (tau, mu) &   quality\_NRMSE \\
\midrule
\endfirsthead
\toprule
                  method & channel &   N &   time\_sec &   mem\_MB &  (tau, mu) &   quality\_NRMSE \\
\midrule
\endhead
\midrule
\multicolumn{7}{r}{\small Continued on next page}
\\\midrule
\endfoot
\bottomrule
\endlastfoot
BIC (actual) &     x,y,z &  1000 &      6.31517 &   13.5474 &    (99, 2) &   1.50297e-05 \\
\midrule
            AMI+FNN &     x &  1000 &     0.22394 &   0.347089 &     (15, 3) &      0.347089 \\
              Cao's &     x &  1000 &     0.185297 &   0.328373 &    (15, 11) &      0.328373 \\
     Unified Embedding Algorithm  &     x &  1000 &     0.253658 &   0.288704 &    (120, 8) &      0.001916 \\
      C--C  &     x &  1000 &      10.0614 &   75.7413 &     (59, 2) &      0.739594 \\
\midrule
            AMI+FNN &     y &  1000 &     0.22394 &   0.347089 &     (16, 4) &      0.347089 \\
              Cao's &     y &  1000 &     0.185297 &   0.328373 &    (15, 11) &      0.328373 \\
     Unified Embedding Algorithm  &     y &  1000 &     0.253658 &   0.288704 &    (120, 8) &      0.001916 \\
      C--C  &     y &  1000 &      10.0614 &   75.7413 &     (49, 2) &       0.44148 \\
\midrule
            AMI+FNN &     z &  1000 &     0.22394 &   0.347089 &     (16, 4) &      0.347089 \\
              Cao's &     z &  1000 &     0.185297 &   0.328373 &    (15, 11) &      0.328373 \\
     Unified Embedding Algorithm  &     z &  1000 &     0.253658 &   0.288704 &    (120, 8) &        0.0019 \\
      C--C  &     z &  1000 &      10.0614 &   75.7413 &     (49, 2) &      0.739594 \\
      \midrule
BIC  &     x,y,z & 12000 &      408.386 &   1510.67 &     (99, 2) &   1.50297e-05 \\
\midrule
            AMI+FNN &     x & 12000 &      1.01517 &    3.52901 &     (16, 4) &      0.347089 \\
              Cao's &     x & 12000 &     0.809524 &    3.12453 &    (15, 11) &      0.328373 \\
     Unified Embedding Algorithm  &     x & 12000 &      1.12243 &    1.27756 &     (14, 2) &       0.00191 \\
      C--C  &     x & 12000 &       408.386 &    1510.67 &     (49, 2) &      0.739594 \\
\midrule
            AMI+FNN &     y & 12000 &       1.01517 &    3.52901 &     (16, 4) &      0.347089 \\
              Cao's &     y & 12000 &      0.809524 &    3.12453 &    (15, 11) &      0.328373 \\
     Unified Embedding Algorithm  &     y & 12000 &       1.12243 &    1.27756 &     (14, 2) &       0.00191 \\
      C--C  &     y & 12000 &        408.386 &     1510.67 &     (49, 2) &      0.44148 \\
\midrule
            AMI+FNN &     z & 12000 &        1.01517 &      3.52901 &      (16, 4) &      0.347089 \\
              Cao's &     z & 12000 &       0.809524 &      3.12453 &     (15, 11) &      0.328373 \\
     Unified Embedding Algorithm  &     z & 12000 &        1.12243 &      1.27756 &       (14, 2) &       0.78979 \\
      C--C  &     z & 12000 &         408.386 &       1510.67 &      (20, 5) &       0.440172 \\
      \midrule
\caption{Benchmark of delay–embedding selectors on synthetic Lorenz data.
Columns: \textit{method} = parameter–selection procedure; \textit{channel} = state variable used (\(x,y,z\));
\(N\) = number of time samples; \textit{time\_sec} = wall-clock runtime per run (seconds);
\textit{mem\_MB} = peak resident memory (MB);
\((\tau,\mu)\) = selected delay interval and number of delays;
\textit{quality\_NRMSE} = normalized RMSE of DMD reconstruction on the holdout set (lower is better).
}
\label{Table:OptimizationDelay} 
\end{slongtable}

%% file: Supp_AlgorithmStructures.tex
\vspace{0.0cm}
\begin{sfigure}[ht]
\begin{center}
    \hspace{-1.3cm}
    \includegraphics[width=1\textwidth, trim={0cm 0cm 0cm 0cm}]{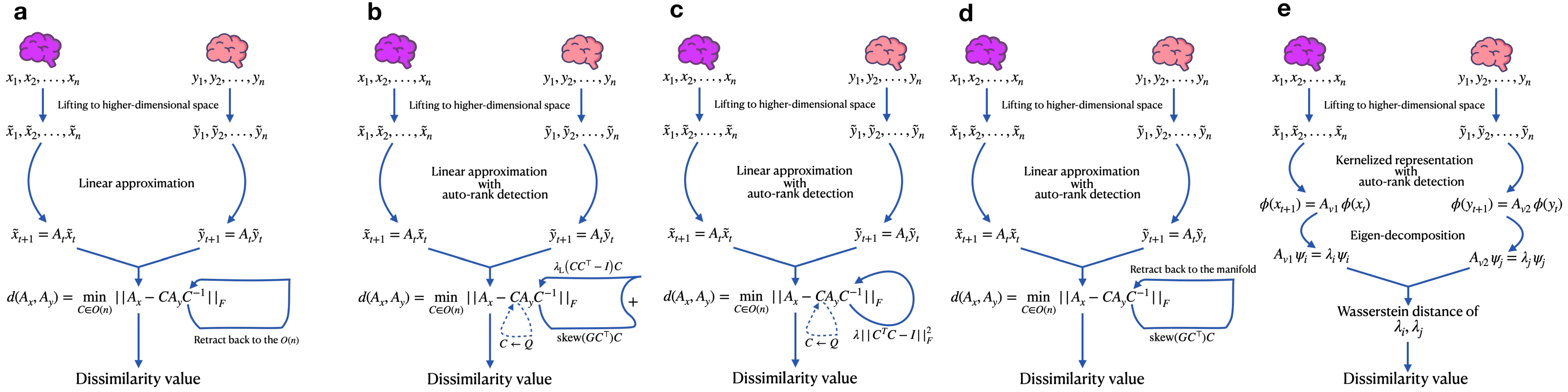}
\end{center}
\vspace{-0.0cm}
\caption{
\textbf{Algorithmic representations of DSA, fastDSA variants, and kwDSA:}
\textbf{(a)} DSA. 
\textbf{(b)} Regularized optimization (RO-fastDSA).
\textbf{(c)} On-manifold Riemannian method (Rim-fastDSA). 
\textbf{(d)} Landing algorithm (Land-fastDSA). 
\textbf{(e)} kwDSA.
}
\label{fig:AlgoritmStructure} 
\end{sfigure}

%% file: fig_maxOrmin.tex
\vspace{0.0cm}
\begin{sfigure}[ht]
\begin{center}
    \hspace{-1.3cm}
    \includegraphics[width=0.32\textwidth, trim={0cm 0cm 0cm 0cm}]{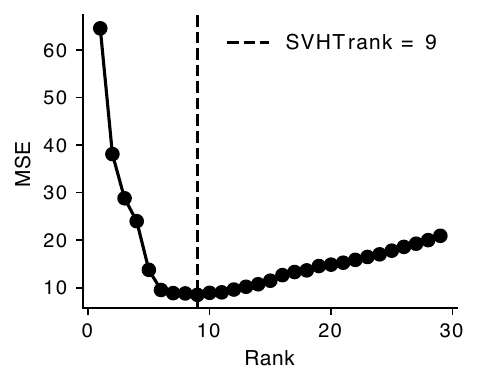}
\end{center}
\vspace{-0.0cm}
\caption{
\textbf{Effect of the choice of the rank on reconstruction quality.}
MSE is computed between the noiseless latent trajectory and the reconstruction obtained by fitting DMD to a Hankel embedding of the noise-corrupted signal (\(n_{\text{delays}}=68\), delay interval \(=1\)). 
The vertical dashed line marks the SVHT-selected rank \(r_{\mathrm{SVHT}}\) estimated from the singular spectrum of the noisy Hankel matrix. The curve typically rises sharply for underestimation (\(r<r_{\mathrm{SVHT}}\)) and flattens for modest overestimation, suggesting that choosing a rank at or slightly above \(r_{\mathrm{SVHT}}\) (e.g., the maximum across paired embeddings) is a safer default with respect to reconstruction error.
}

\label{fig:minOrmax} 
\end{sfigure}


%% file: fig_Landscape_transformation.tex
\vspace{0.0cm}
\begin{sfigure}[ht]
\begin{center}
    \hspace{-1.3cm}
    \includegraphics[width=0.6\textwidth, trim={0cm 0cm 0cm 0cm}]{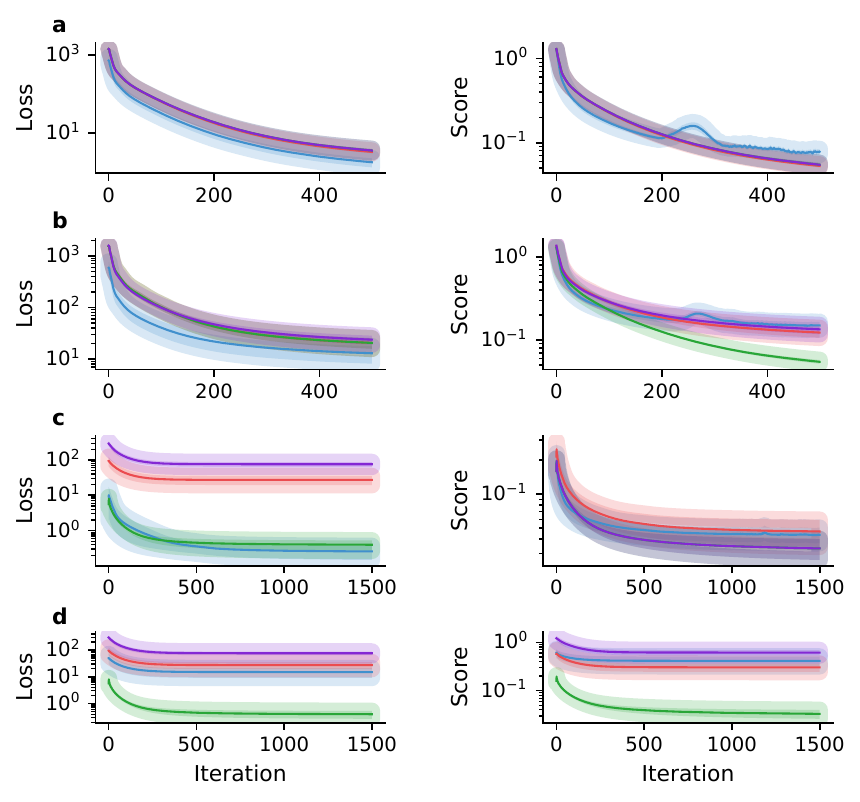}
\end{center}
\vspace{-0.0cm}
\caption{\textbf{Analysis of different deformation and transformation scenarios :} 
Comparison of the loss (left) and difference scores (center) in deformation experiments between DSA and fastDSA variants across the optimization process (500 iterations). 
\textbf{a}: $\beta = 1$. 
\textbf{b}: $\beta = 4$.
Comparison of the loss (left) and difference scores (center) in the transformation experiment between DSA and fastDSA variant across iterations. \textbf{c}: $\alpha = 0.2$. 
\textbf{d}: $\alpha = 0.9$. 
The Shaded regions in the line plots indicate ±1 standard error of the mean.}
\label{fig:FigureLandscapetransformation} 
\end{sfigure}

%% file: fig_supp_kwDSA.tex
\vspace{0.0cm}
\begin{sfigure}[ht]
\begin{center}
    \hspace{-1.3cm}
    \includegraphics[width=1.02\textwidth, trim={1cm 0cm 0cm 0cm}]{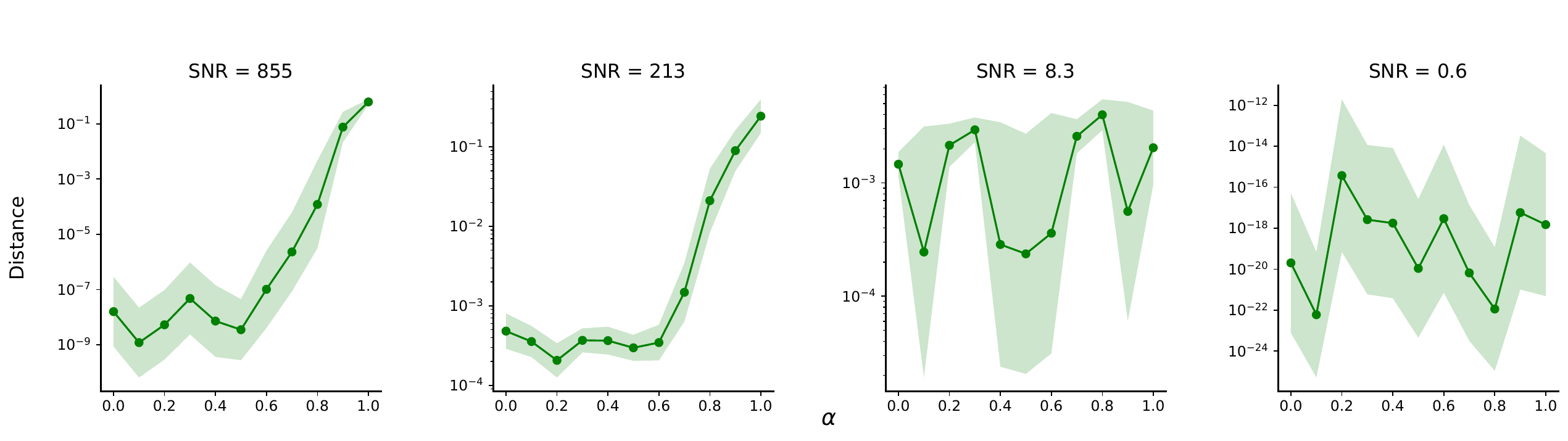}
\end{center}
\vspace{-0.0cm}
\caption{
\textbf{Log-scale plot of \autoref{fig:Figure5_noisy}a:}
Similarity (distance) as a function of the parameter \(\alpha\) estimated by kwDSA for four different signal-to-noise ratios (SNR = 855, 213, 8.3, 0.6) on a logarithmic scale. The plot shows the same data as the kwDSA curves in \autoref{fig:Figure5_noisy}a, but with a logarithmic y-axis to better visualize performance differences, especially at lower distances.
Shaded regions in the line plots indicate ±1 standard error of the mean across 10 runs. 
}
\label{fig:FigurekwDSA_log} 
\end{sfigure}

%% file: FigSupp_twoSystem_stat.tex
\vspace{0.0cm}
\begin{sfigure}[ht]
\begin{center}
    \hspace{-1.3cm}
    \includegraphics[width=1\textwidth, trim={0cm 0cm 0cm 0cm}]{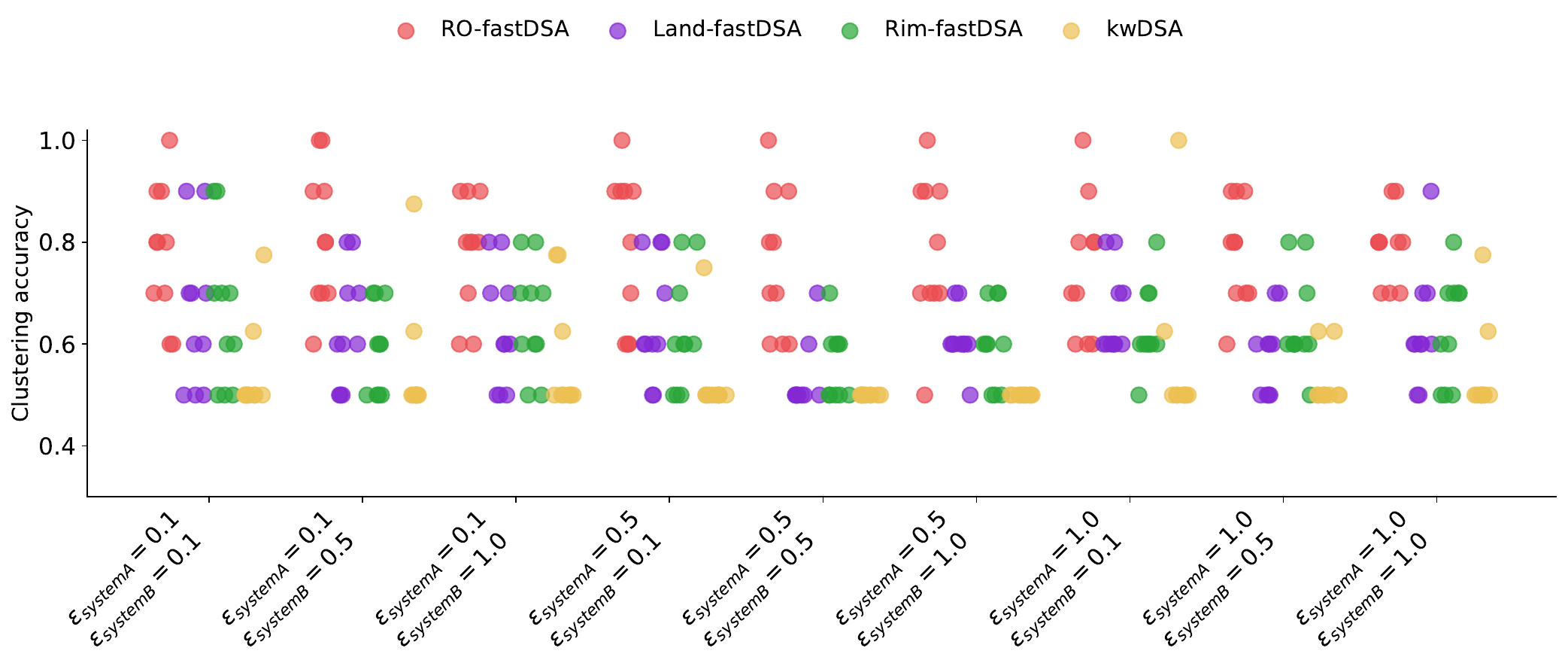}
\end{center}
\vspace{-0.0cm}
\caption{
\textbf{Comparing fastDSA and kwDSA:}
Comparing clustering accuracy between fastDSA and kwDSA across different choices of the parameter $\epsilon$, using $k$-means clustering with $n_{\mathrm{clusters}} = 2$, a single random initialization, and a fixed random seed to ensure reproducibility. Across combinations of $\epsilon$ for the two systems, the family of fastDSA methods consistently yields higher clustering accuracy than kwDSA, indicating greater sensitivity to subtle changes in the governing dynamics because fastDSA compares \emph{both} eigenvalues and eigenvectors, whereas kwDSA focuses only on eigenvalue distributions.
}
\label{fig:FiguretwoSystemAccuracyStat} 
\end{sfigure}

%% file: fig_delayOptimization.tex
\vspace{0.0cm}
\begin{sfigure}[ht]
\begin{center}
    \hspace{-1.3cm}
    \includegraphics[width=1\textwidth, trim={0cm 0cm 0cm 0cm}]{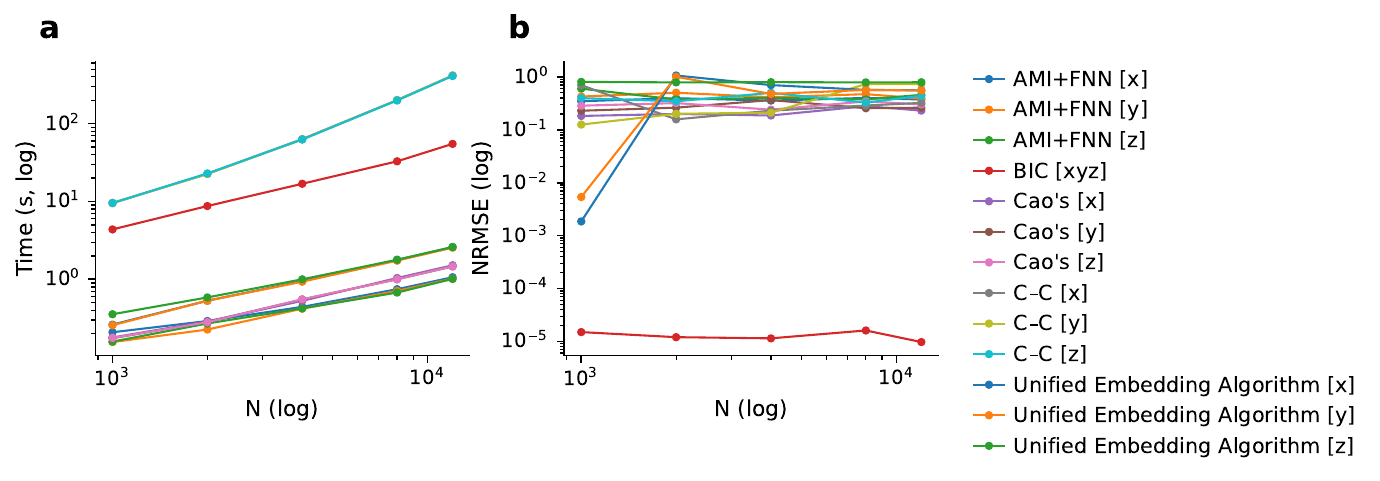}
\end{center}
\vspace{-0.0cm}
\caption{\textbf{Scalability and reconstruction accuracy of delay–embedding selectors on synthetic Lorenz-based data.}
\textbf{(a)} Clock runtime versus sample size \(N\) on log–log axes for five embedding–parameter selectors (discussed in section \nameref{sec:supp-embed}). BIC is evaluated jointly across all channels (single curve), whereas the other methods are applied per channel; curves are labeled as \(\textit{method}[\textit{channel}]\) with \(\textit{channel}\in\{x,y,z\}\).
\textbf{(b)} Normalized reconstruction error (NRMSE) versus \(N\) on log–log axes (lower is better), computed by fitting DMD using the embedding parameters chosen by each method and comparing the reconstruction to the original synthetic Lorenz data. Across all sample sizes, BIC achieves the best reconstruction accuracy but at substantially higher computational cost, whereas faster heuristics (e.g., projection+ACF, AMI+FNN, Cao’s method) scale more favorably but sacrifice predictive fidelity. No method exhibits uniformly strong performance across both speed and accuracy, underscoring that reliable and scalable delay–embedding selection remains an open challenge.
}

\label{fig:delayOptimization} 
\end{sfigure}
